\documentclass[10pt]{article}
\usepackage[letterpaper,margin=1in]{geometry}
\usepackage{microtype}
\usepackage{graphicx}
\usepackage{booktabs}
\usepackage{amsmath,amssymb}
\usepackage{enumitem}
\usepackage{xcolor}
\usepackage{url}
\usepackage{hyperref}
\usepackage[nameinlink,capitalize]{cleveref}
\usepackage{authblk}
\usepackage[numbers,sort&compress]{natbib}
\usepackage{caption}
\usepackage{subcaption}
\usepackage{float}
\usepackage{tcolorbox}

\usepackage[autostyle, english = american]{csquotes}
\MakeOuterQuote{"}

\usepackage{fancyvrb}
\usepackage{fvextra}

\hypersetup{
  colorlinks=true,
  linkcolor=black,
  citecolor=black,
  urlcolor=black
}
\DefineVerbatimEnvironment{promptbox}{Verbatim}{
  fontsize=\footnotesize,
  frame=single,
  framerule=0.5pt,
  rulecolor=\color{gray},
  xleftmargin=1em,
  xrightmargin=1em,
  breaklines=true,
  breakanywhere=true
}

\title{Embedded AI Companion System on Edge Devices}
\author[1]{Rahul Gupta\thanks{\texttt{rgupta2718@gmail.com}}}
\author[1,2]{Stephen Hsu}
\affil[1]{Superfocus AI}
\affil[2]{Michigan State University}
\date{\today}

\begin{document}
\maketitle

\begin{abstract}
Computational resource constraints on edge devices make it difficult to develop a fully embedded AI companion system with a satisfactory user experience. AI companion and memory systems detailed in existing literature cannot be directly used in such an environment due to lack of compute resources and latency concerns. In this paper, we propose a memory paradigm that alternates between active and inactive phases: during phases of user activity, the system performs low-latency, real-time dialog using lightweight retrieval over existing memories and context; whereas during phases of user inactivity, it conducts more computationally intensive extraction, consolidation, and maintenance of memories across full conversation sessions. This design minimizes latency while maintaining long-term personalization under the tight constraints of embedded hardware. We also introduce an AI Companion benchmark designed to holistically evaluate the AI Companion across both its conversational quality and memory capabilities. In our experiments, we found that our system (using a very weak model: \texttt{Qwen2.5-7B-Instruct} quantized int4) outperforms the equivalent raw LLM without memory across most metrics, and performs comparably to GPT-3.5 with 16k context window.
\end{abstract}

\section{Introduction}
There are many potential applications for AI companions or assistants, ranging from entertainment to psychological well-being and educational support~\cite{Maples2025-nz, SKJUVE2021102601} (e.g., language learning~\cite{DU2024100230}, social skill development, etc.). In this paper we focus specifically on an AI companion toy designed for children.

It is advantageous to run a companion system entirely offline, embedded directly on an edge device, for many reasons. First, it considerably reduces privacy concerns. Since all processing occurs on device, no personal user data (in this case, children’s conversational data) needs to be transmitted to external servers. Second, it can lower costs. A toy manufacturer may incur a one-time expense for the hardware rather than recurring fees for cloud inference or third-party API usage. This in turn enables more transparent, one-time pricing for consumers, rather than subscription models tied to server use. Finally, it can operate in settings with limited or unreliable internet connectivity, allowing for wider adoption and consistent functionality.

Prior work has demonstrated that companion systems need to be able to handle long-term memory across a series of conversations and personalize towards users accordingly~\cite{li2025helloagainllmpoweredpersonalized}. However, these works typically assume powerful cloud-based LLMs, relaxed latency budgets, or the availability of large context windows (\cref{sec:related}).

Our setting departs significantly from these assumptions. Due to computational constraints, edge devices can host only small, quantized models. These models are less intelligent and have smaller context windows than large cloud-based LLMs. Additionally, due to these computational constraints, the response latency can be high. This is especially true if given long input context, therefore forcing us to even further limit our already small context window for the sake of low latency. Moreover, high latency makes multiple sequential LLM requests infeasible, such as for real-time memory extraction during user interaction. These constraints create a mismatch with prior formulations of long-term memory pipelines and motivate the need for a new paradigm that preserves real-time responsiveness while still learning over extended conversations.  

Prior works have developed benchmarks intended to test AI systems' memory capabilities~\cite{maharana2024evaluatinglongtermconversationalmemory}, and similarly there are benchmarks intended to evaluate AI systems on other aspects of conversational quality~\cite{cheng2025autopalautonomousadaptationusers}. However, there is no single benchmark intended to evaluate all these relevant aspects for an AI companion system. Consequently, to comprehensively evaluate an AI system using prior benchmarks, one would have to run multiple separate benchmarks to evaluate each aspect. Furthermore, many of these existing evaluation methods require some level of human feedback~\cite{cheng2025autopalautonomousadaptationusers}, or rely on a curated task-specific set of example benchmark inputs or conversations~\cite{maharana2024evaluatinglongtermconversationalmemory}.

We propose a fully automated benchmark using large LLMs for user simulation and as judges for evaluation. The benchmark utilizes multiple evaluation criteria to holistically evaluate various aspects of the system's performance, including the quality of its extracted memories, quality of the conversation, ability to retrieve and utilize those memories for targeted questions, as well as its displayed ability to personalize to the user in a more natural conversation. Our context engineering tailors the benchmark towards our AI toy use case, but this benchmark framework could easily be adapted towards other AI companion use cases with simple prompt changes.

 In this paper, to address the challenges of our assumed environment, we introduce a memory framework that separates processing into phases of user activity (real-time, low-latency) and inactivity (deferred, higher-cost). We use our benchmark to show that our system outperforms a raw Qwen model without memory, thereby showing the viability of running a fully embedded AI companion system on edge devices. 
 
 In this paper, we summarize the key contributions as follows:
 \begin{itemize}[leftmargin=*]
 \item We illustrate specific challenges for embedded AI environments that cause many common memory system frameworks to be infeasible. We then introduce a latency-oriented AI companion system designed for embedded AI based on phases of user activity. 
 \item We introduce a new fully automated, holistic benchmark tailored towards our AI toy use case. However, we hold that, with only prompt changes, the same benchmark framework could be utilized for other AI companion use cases as well.
\item Using our benchmark, we show the viability of running an embedded AI companion system, as well as illustrate areas where it struggles compared to an unrestricted system using larger models.
\end{itemize}

\section{Related Work}
\label{sec:related}
\paragraph{Conversational, personalized, LLM-based, Long-term Memory Systems.} 
Many modern LLM-based systems utilize memory components for conversation with users. There is growing interest in personalized AI systems, which need to (a) keep track of user information and events mentioned over the course of the conversation and (b) use the previously mentioned information to better craft future responses correctly and tailor them towards the user. Modern systems in the literature typically decompose into an extraction, update/refinement, and retrieval action~\cite{tan2025prospectretrospectreflectivememory, zhong2023memorybankenhancinglargelanguage, ong2025lifelongdialogueagentstimelinebased, chhikara2025mem0buildingproductionreadyai}, though exact implementations vary. Extraction distills memories and potentially also a user persona~\cite{li2025helloagainllmpoweredpersonalized} from dialogue; update integrates them into a store (database~\cite{li2025helloagainllmpoweredpersonalized}, knowledge graph~\cite{ong2025lifelongdialogueagentstimelinebased}, etc.), handling conflicts and forgetting; retrieval selects relevant memory items for subsequent turns. 

Performing additional LLM inference requests besides just the single main response generation LLM request (that generates the response to the user utterance), either every turn or set amount of turns, can lead to better quality responses. \cite{ong2025lifelongdialogueagentstimelinebased} uses an LLM request to perform refinement of memories when generating each new user response. \cite{li2025helloagainllmpoweredpersonalized} uses an LLM request for extraction of persona information. \cite{Perera_2024} uses an LLM request for summarizing ongoing context in order to maintain the length of the current ongoing short term conversation. 

In contrast to system designs which perform extraction or refinement every turn or set number of turns, some systems such as RMM~\cite{tan2025prospectretrospectreflectivememory} perform extraction at the end of a dialogue session, in order to improve memory quality by performing extraction on less fragmented finished dialogue sessions. RMM~\cite{tan2025prospectretrospectreflectivememory} also introduces a reranker that is fine-tuned during the user conversation, to improve memory retrieval over time. 

As quality is typically the focus, many of these systems use larger sized or online API-based LLMs such as models from the GPT/Gemini families, etc. \cite{ong2025lifelongdialogueagentstimelinebased, tan2025prospectretrospectreflectivememory}. However, some papers also demonstrate performance with smaller open source LLMs:  \cite{li2025helloagainllmpoweredpersonalized} also demonstrates performance with ChatGLM (6B)~\cite{zeng2023glm130bopenbilingualpretrained}, and \cite{zhong2023memorybankenhancinglargelanguage} also demonstrates performance with BELLE (7B)~\cite{BELLE}. Some systems like Mem0~\cite{chhikara2025mem0buildingproductionreadyai} also explicitly focus on lowering latency, through architecture design and asynchronism. 

Many of the memory systems listed above don't explicitly mention the behavior of their systems under the situation where the ongoing short term conversation reaches a context window that is `too long' (either with regards to latency, or with regards to hitting the max context window for models with smaller context windows i.e. 30k for Qwen 7B). For systems exclusively using large models, this may not be a realistic issue, but even many papers that otherwise mention usage with a small context window model don't explicitly mention their behavior in such a situation~\cite{li2025helloagainllmpoweredpersonalized}, possibly because of an assumption that sessions won't last that long. However, notable exceptions include \cite{packer2024memgptllmsoperatingsystems} which utilize LLMs function calls to dynamically allow it to manage its own context window, and systems like \cite{chhikara2025mem0buildingproductionreadyai} which utilize an LLM for summarization of ongoing conversation in conjunction with a sliding message window that only looks at some most recent amount of messages.  

Our environment is most similar to~\cite{chhikara2025mem0buildingproductionreadyai} which also focuses on latency, and has some mention of context management. However, due to the computational limitations of our edge device, we are forced to impose additional limitations on our architecture and model choices.

\paragraph{Benchmarks and Synthetic Evaluation.} Benchmarks such as RULER~\cite{hsieh2024rulerwhatsrealcontext} are most commonly used to evaluate LLM memory capabilities. These types of benchmarks are purely memory-focused and are designed mainly to test an LLM's working memory by giving it large chunks of context at once. However, single isolated tasks are not a good representation of how a conversational model learns about a user over time. Consequently, benchmarks like LoCoMo~\cite{maharana2024evaluatinglongtermconversationalmemory} were developed to better evaluate a conversational AI system's ability to extract and remember information across sessions for a well-defined user. However, since LoCoMo's~\cite{maharana2024evaluatinglongtermconversationalmemory} tasks give you the partial conversation history, it cannot fully test an AI companion's capability to drive a conversation and its ability to elicit information from the user. For example, an AI companion system might be good at directly responding to a user, but not ask the user for additional details to learn more about them. Such behavior is not evaluated by LoCoMo~\cite{maharana2024evaluatinglongtermconversationalmemory}, since the overall course of the conversation is already determined by the ground truth conversation, rather than being controlled by the actual AI companion system. However, similar to LoCoMo~\cite{maharana2024evaluatinglongtermconversationalmemory}, we do use a LLM to synthetically play the role of a user, given a well defined persona. Similarly, we also use LLMs as judges to evaluate responses on our criteria, a practice established by prior papers.~\cite{tan2025prospectretrospectreflectivememory, chhikara2025mem0buildingproductionreadyai}.

Conversational AI papers focus their evaluation more on criteria related to conversational quality, rather than testing a specific QA or generating a response given partial history. Like the memory benchmarks, these evaluation criteria instead require the AI system to have a full conversation, after which the AI system's performance across the conversation as a whole is evaluated. For example, AutoPal~\cite{cheng2025autopalautonomousadaptationusers}, evaluates the agent on criteria such as naturalness, affinity, and personalization. We also adopt some of these criteria into our benchmark, although whereas these criteria are typically evaluated by humans, we synthetically evaluate them using a larger model.

\section{Problem Formulation}
`Edge' device is a relative term that captures a wide range of devices. In the field of robotics, even an RTX 4090 or 3090 can be considered an edge device~\cite{park2025specedgescalableedgeassistedserving}. Here we define edge devices to have much weaker capabilities. Specifically, we test using the NVIDIA Jetson Orin Nano Super 8GB~\cite{NVIDIA_JetsonOrinNanoSuper}, enforcing the following set of assumptions on our system:
\begin{enumerate}[leftmargin=*]
  \item \textbf{Memory constraints:} Under the assumption of only 8GB of VRAM, devices can fit only small, quantized models. This becomes even more challenging if developing a full pipeline with additional models, such as a STT and TTS. To balance these challenges while also maximizing the intelligence of our LLM model, we opt to use a 7B parameter model with int4 quantization.
  \item \textbf{Compute constraints:} Edge devices have orders of magnitude less compute available than cloud devices. Increasing context window or running parallel requests will quickly hit memory limits and increase latency.
  \item \textbf{Real-time constraints:} Responsiveness matters for a natural companion UX. Studies show that human conversations with faster response times are rated as more enjoyable and related to increased feelings of social connection ~\cite{fast_response_social_conn}. Average human response time in English is 236 ms with a standard deviation of 519 ms~\cite{human_response_time}. Therefore, we aim to minimize perceived latency wherever possible, unless at a very high cost to quality. Any such exceptions will be clearly noted in the paper.
  \item \textbf{Limited Context Window:} On such devices, we cannot make use of the full context window even for these smaller models. \cref{fig:latency_ttft} illustrates the average \emph{time to first token} (TTFT) latency for Qwen 7B. We attempt to replicate realistic conditions in our latency testing, for implementation details refer to Appendix \cref{appendix_latency}. Beyond 10k tokens, llama-server begins to run out of memory due to the computational resource limits of the Jetson. However, long before it reaches that point, smaller context windows would be necessary to maintain reasonable latencies. To achieve a goal of even $\sim$5s, context windows smaller than $\sim2500$ tokens would be necessary. Human-like latencies of $\sim2$s would require limits of $\sim$1000 tokens. Similarly, \cref{fig:latency_tps} shows that a similar context window threshold would be needed to maintain a reasonable tokens/sec output as well.

\begin{figure}[H]
    \centering
    
    \begin{minipage}[t]{0.48\textwidth}
        \centering
        \includegraphics[width=\textwidth]{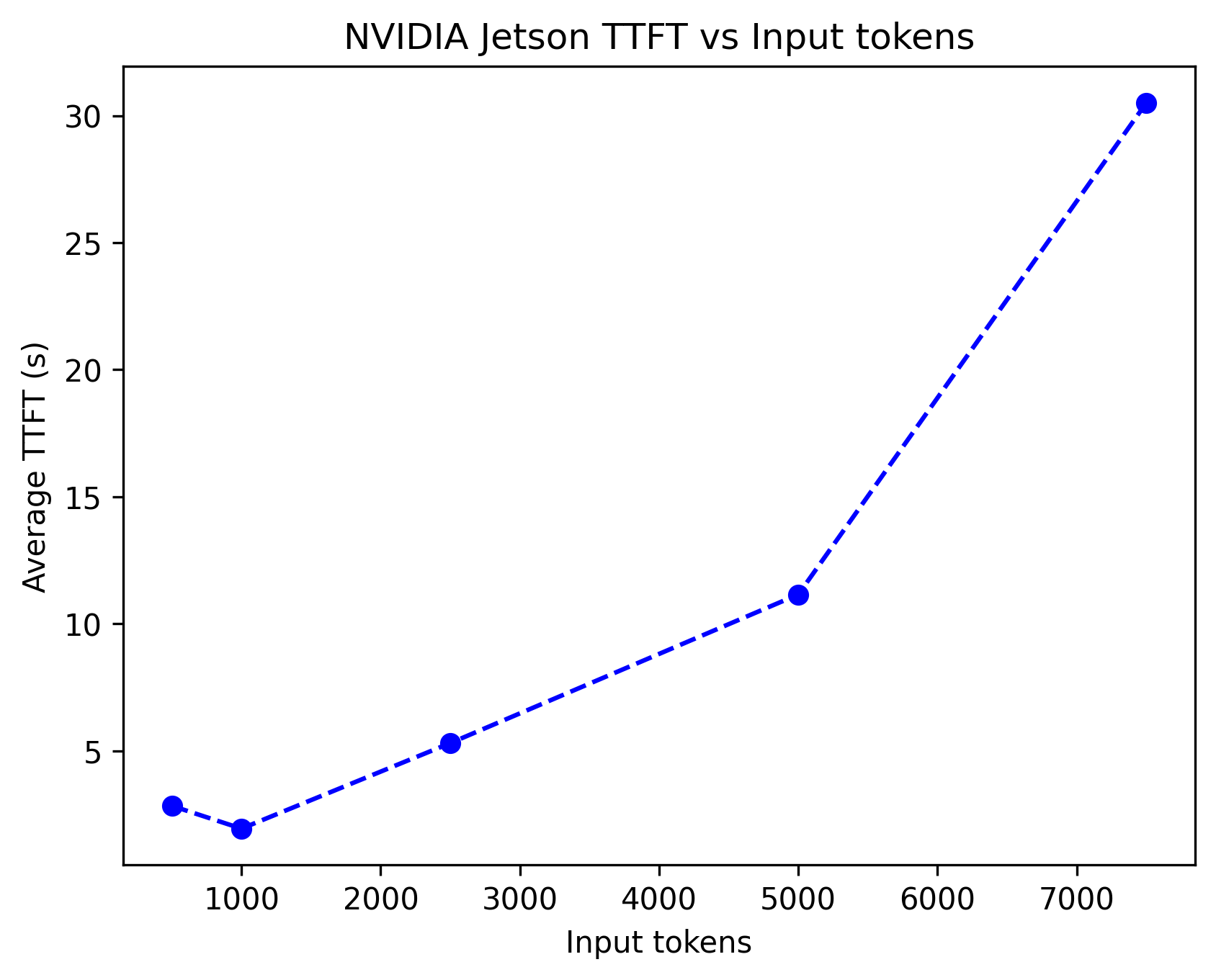}
        \caption{Average TTFT across input token sizes on NVIDIA Jetson. We use \texttt{Qwen2.5-7B-Instruct} (int4) GGUF~\cite{qwen2.5, qwen2} using llama.cpp~\cite{llama_cpp} for requests with varying input token size, taken over 5 trials.}
        \label{fig:latency_ttft}
    \end{minipage}
    \hfill
    \begin{minipage}[t]{0.48\textwidth}
        \centering
        \includegraphics[width=\textwidth]{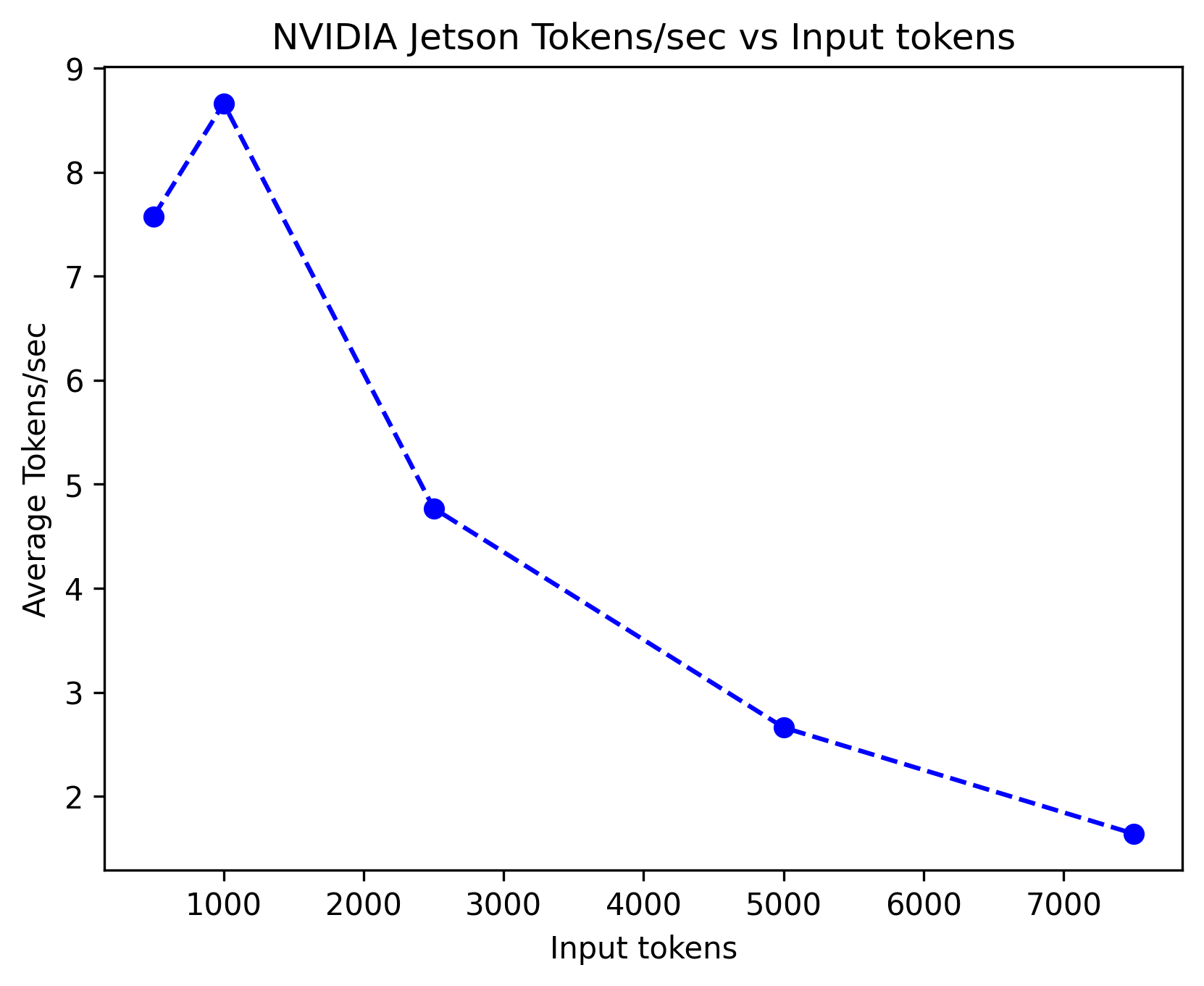}
        \caption{Average tokens/sec across input token sizes on NVIDIA Jetson. Same settings are used.}
        \label{fig:latency_tps}
    \end{minipage}

\end{figure}
  
\end{enumerate}
Given the above set of assumptions, the system could run on even cheaper hardware than the Jetson, as a 7B int4 model can even be run on a good quality cpu. Similarly, there are some chips which advertise npus that can run these models.

\section{Method}
\label{sec:method}
\begin{figure}
    \centering
    \includegraphics[width=0.6\textwidth]{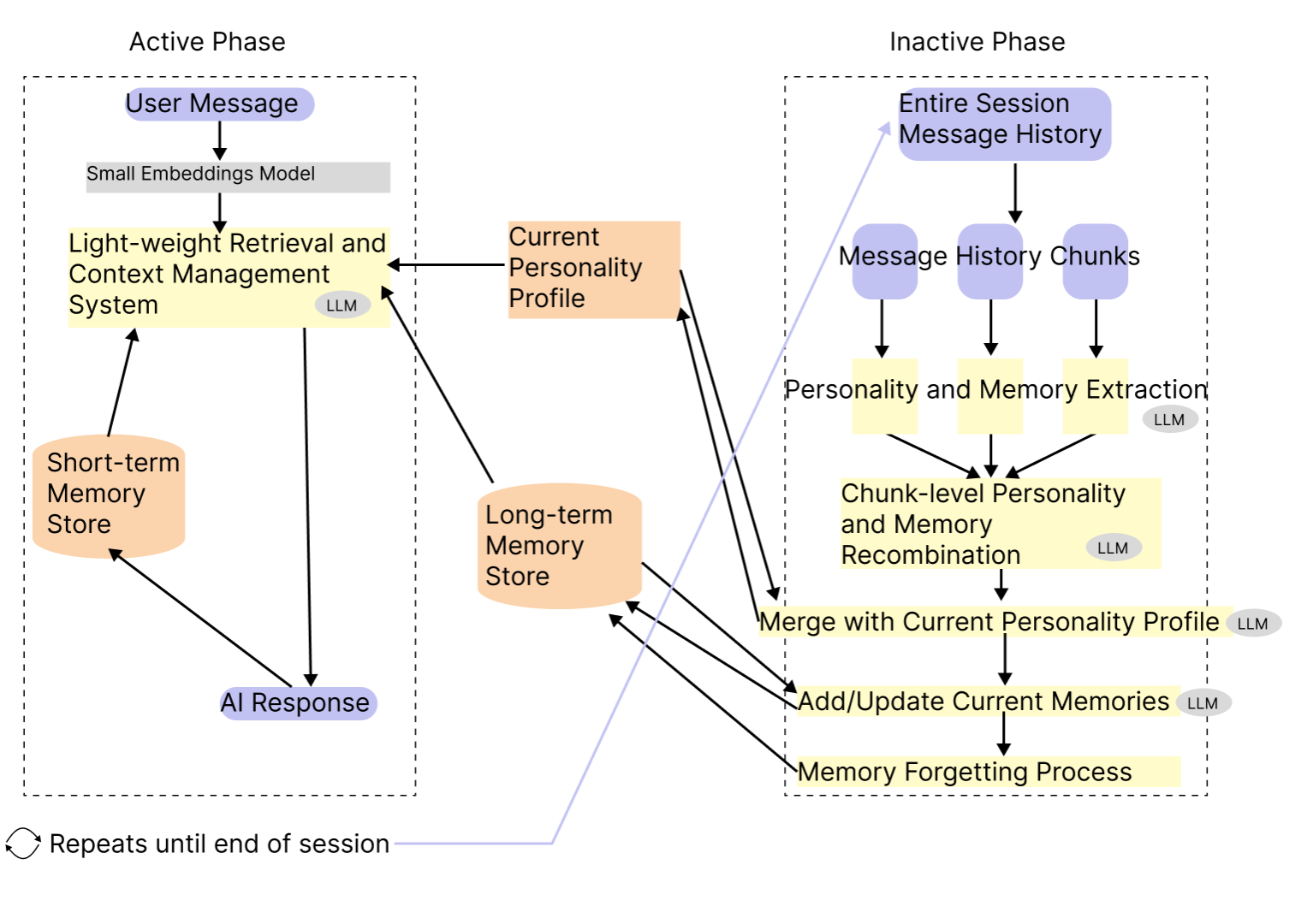}
    \caption{Overview of System Architecture.}
    \label{fig:system_arch}
\end{figure}

\subsection{Overview}
\cref{fig:system_arch} shows an overview of our system. Our architecture utilizes an active--inactive memory paradigm. During phases of user activity, we perform low-cost memory retrieval of any relevant long term and short term memories. We store query-response pairs from the current ongoing session, called \emph{short term memories}, into a simple short term memory store. Then during phases of user inactivity, we perform the more computationally intensive task of extracting \emph{long term memories} and user personality profiles from the current session of conversation. Long sessions may need to be split into chunks, from each of which memories and user profiles will be extracted and then merged into a final set of memories and a final user profile. These session level memories and user profile are then consolidated with the existing set of memories and user profile. Individual memories then go through an additional refinement step where similar existing memories may be merged or overwritten. For all parts of our system (and relevant baselines), we run \texttt{Qwen2.5-7B-Instruct}~\cite{qwen2.5, qwen2} with int4 gguf quantization on llama.cpp (commit: 9f052478c)~\cite{llama_cpp}.

\subsection{Retrieval and Generation at Inference Time (Active)}
\label{sec:active}
We define a conversation \emph{session} as a single continuous (i.e. without long breaks) dialogue between the user and the AI system. Each session is composed of multiple conversation \emph{turns}, or individual back-and-forths between the user and the AI. At each turn $i$ the user's current utterance, or \emph{query} $q$ is embedded to $e_q$. For the embedding model, gte-base-en-v1.5~\cite{zhang2024mgte, li2023gte} is used. Cosine similarity search is performed over stored embeddings to retrieve top-$k_{\text{long}}$ long term memories $\mathcal{M}_{\text{long}}$ and top-$k_{\text{short}}$ short term memories $\mathcal{M}_{\text{short}}$. Of the retrieved memories, only those that pass a fixed retrieval similarity threshold, $S_\text{min}$, are used. This is to prevent the response generation model from potentially using irrelevant memories in its response. To keep message history from getting too long, we keep a sliding window of the $W_\text{slide}$ most recent messages in context. Therefore we depend on short term memories to provide the model with relevant context from outside the sliding window. An individual short-term memory can essentially be thought of as a query-response pair $(q,r)$ from the past dialogue in the current conversation session. For relevant short term memories, the immediate $k_\text{surrounding}$ query-response pairs right before and after the relevant short term memory are also retrieved to provide additional context.

Query $q$, all retrieved memories ($\mathcal{M}_{\text{long}}$ and $\mathcal{M}_{\text{short}}$), sliding context window messages and user profile $p$ are formatted into the response generation system prompt and used to generate the AI system response, $r$. After generating the response, the current turn is also stored in short term memory in the form $(i, q, e_q, r)$, where $i$ is the index of the memory in the current conversation.

\subsection{Extraction and Update (Inactive)}
\label{sec:inactive}
After detecting user inactivity via lack of a new user response within some time threshold, $T_\text{inactivity}$, we enter the inactive phase (marking the end of the current conversation session). Since we are less latency sensitive in this phase, we could load in a more powerful model in this phase if our device allowed (or if we were using something less powerful during the active phase, we could then load in the 7B model for this phase). However, for the Jetson, we use the same quantized Qwen model for both phases. 

During the inactivity phase, we take the entire conversation history from the current finished session, and split it into chunks of $c_\text{chunk}$ tokens. Even though we are less latency sensitive in this phase, we still want it to finish within a reasonable time-frame, ideally before the user starts using the AI toy again. We then extract a \emph{user profile} $p$ and long-term memories $\mathcal{M}_{\text{long}}$ from each chunk. The user profile contains (i.) some set of core user characteristics that we think should always be inserted into the response generation prompt (name, age, gender) and (ii.) a personality profile of the user, inspired by papers like LD-Agent~\cite{li2025helloagainllmpoweredpersonalized}, which has both a high level one to two sentence personality summary, as well as attempts to classify the user on the Meyers Briggs~\cite{myersbriggs_overview_2025} and Big Five~\cite{bigfive} personality scales as more concrete measures. We also extract supporting dialogue examples of these personality qualities to help consolidate old and new personality profiles.

For chunks after the first chunk, we also include the last $k_\text{overlap}$ messages, as well as a LLM generated summary of the past chunk, to help preserve context across chunks when generating extractions.

Our next step is to use an LLM to merge the chunk-level extracted profiles and memories across the given session into a final consolidated profile and set of memories for the session. The session-level profile then once again goes through a merge with the previously remembered user profile and the result becomes the new user profile.

We experimented with a memory post-processing step where each individual new memory would go through additional post-processing to make memories easier for the model to understand (mainly with regards to dates and times). However, we were not satisfied with the 7B model's ability to meaningfully perform this post-processing so we ultimately removed this step. We would likely reincorporate this step if we were able to use a more powerful model. Refer to~\cref{appendix_post_processing} for more details.

The session-level memories then go through a more complex `refinement/update' process for consolidation. Similar to RMM~\cite{tan2025prospectretrospectreflectivememory}, we go through each potential new memory one at a time, compare to similar existing memories and use an LLM to decide whether to overwrite/merge with an existing memory, or to be added as a separate memory entirely. For each memory, we store $(m_{\text{long}}, e_m, T_\text{insert}, T_\text{last\_use}, S)$ where $m_{\text{long}}$ is the textual memory representation, $e_m$ is the embedding, $T_\text{insert}$ is the insert time of the memory (current time at time of extraction), $T_\text{last\_use}$ is the last use time of the memory (initially same as $T_\text{insert}$), and $S$ is the memory strength (explained later in \cref{sec:forgetting}). 

\subsection{Forgetting (Inactive)}
\label{sec:forgetting}
Both for the sake of retrieving good memories and reducing database retrieval latency, it is important for us not to overpopulate our long term memory store. Consequently we adopt the method for forgetting memories as described in MemoryBank~\cite{zhong2023memorybankenhancinglargelanguage}. The MemoryBank paper describes using the formula $R = e^{-t/S}$, inspired by the Ebbinghaus forgetting curve~\cite{ebbinghaus1964memory}, where $R$ is the memory retention; $t$ is the time since learning the information but they reset it to 0 each time a memory is used (for our purposes we treat $t$ as the days since last use, calculated using $T_\text{last\_use}$ and $T_\text{insert}$); and $S$ is the memory strength, initialized to 1 and incremented each time a memory is used~\cite{zhong2023memorybankenhancinglargelanguage}. We do the same: whenever a long term memory is retrieved, we increase its strength $S$ by 1 and update its time of last use $T_\text{last\_use}$. During the inactivity phase, after extraction, we also perform a forgetting action, where we drop memories whose retention value $R$ falls below some retention threshold $R_\text{min}$. 
With that said, we include this mainly just for the sake of presenting a production ready-system. In our tests, we didn't simulate the passage of time, so our evaluation scores aren't affected by our forgetting mechanics.

\subsection{JSON validation and retries (Inactive)}
\label{sec:json}
Various steps during our inactivity phase depend on loading valid JSON output of a specific schema. Our quantized Qwen 7B model will sometimes give invalid JSON. If we receive invalid JSON, we will first attempt a `soft' JSON retry, where we will give the LLM the invalid JSON and ask it to correct it. If this fails, then we will perform a full retry where we will rerun whichever prompt yielded the invalid JSON in question. We attempt the `soft' retry first, since many of our extraction prompts take large token inputs (such as the large chunks of the session message history), significantly decreasing the latency of the `soft' retry. 

\subsection{Greeting Phase (Active)}
\label{sec:greeting}
Also of note is that the first time the user talks to our system, we go through a special `greeting phase'. The goal of the greeting phase is to immediately extract a few `core' user characteristics like the users' age, name, gender that are always relevant for the response generation model. Upon extraction, these items are immediately put into the current user profile $p$ so they will remain in the response generation prompt even after the greeting phase conversation leaves the message history sliding window. 

In the priority order of name then age then gender, we attempt to immediately extract these pieces of information. To avoid making multiple LLM requests, we attempt to use a single LLM request to both extract the piece of information (if it exists) and simultaneously respond accordingly (ask for the information if it doesn't exist, otherwise converse with the user normally).
Example responses might look like \texttt{\detokenize{{"name": "John", "assistant_response": "response to the user"}}} or \verb|{"name": "null", "assistant_response":| \\ \verb|"response asking for the user's name"}|. On a null value, we try again. On a non-null value, we move on to the next information item. However, we recognize that our small and quantized model struggles to reason over this and our system sometimes ends up asking for items a 2nd time after they've already been provided.

\section{Experimental Setup and Results}
\label{sec:benchmark}
\subsection{Overview}
We use GPT-5~\cite{OpenAI_GPT5_2025} to synthetically generate a comprehensive user \emph{characterization}, including details such as their name, age, gender, likes/dislikes, personality profile, etc. We use Anthropic's Claude Sonnet 4.5~\cite{anthropic2025_claudesonnet4_5} to play the role of the user based on the given characterization. In this way, we simulate long-term conversations with our system across multiple sessions, each consisting of a number of "turns" between the simulated user and AI companion (running a small LLM, Qwen, on Jetson). Where each "session" is intended to represent a single 1-3 hour conversation sitting between the user and the AI, after which the user leaves for a more extended period of time (i.e. until the next day).  After each session, Claude generates a summary of said session. All Claude-generated summaries are also used to generate subsequent sessions. The AI companion responses and extracted memories are then evaluated across various categories, which serve as a holistic evaluation of the AI Companion system, assessing the performance of various sub-parts within the larger system. We utilize OpenAI's GPT-5 as the model responsible for grading and evaluation. We evaluate our system and comparison baselines along 3 dimensions: (i.) \textbf{Conversation Quality} (overall quality of AI responses across the conversation), (ii.) \textbf{Generated QA} (quality of AI answers to specific questions we generate based on the conversation), (iii.) \textbf{Personalization} (amount of demonstrated examples of personalization during a short new conversation session). We also evaluate our system (as its the only one with memory capabilities) on an additional dimension: \textbf{Extraction Quality} (quality of memory extraction capabilities).

\subsection{Comparison Baselines}
For comparison baselines, we firstly compare to the same quantized Qwen 7B model with no memory, instead using a sliding message history window of 30k tokens (as the context window is limited to 32k tokens). While such a device can't be feasibly run on the edge devices in question, we still run comparisons on an A100 GPU just to showcase the benefits of our memory system versus having no memory system. Secondly, we compare to raw GPT-3.5 Turbo~\cite{gpt35turbo_2025} with a 15.5k sliding window (as the context window is 16k tokens). This is intended to show a comparison against a smarter more intelligent model that still has a small context window. Finally, we also compare to the raw GPT-5 model with the entire message history in context. This is to illustrate the limits of small models and current embedded edge AI performance compared to large online models.

For all provided results below, we run 5 user characterizations for 100k tokens total, split across 10 sessions, 10k tokens each, for our system and the two baselines (same characterizations used).

\subsection{Conversation Quality}
\label{subsec:conv_quality}
This evaluation criteria focuses more on subjective measures inspired by AutoPal's~\cite{cheng2025autopalautonomousadaptationusers} naturalness, affinity, and personalization evaluation criteria. Although AutoPal~\cite{cheng2025autopalautonomousadaptationusers} utilized human evaluation, we use GPT-5 to automate the evaluation process. We give GPT-5 the whole conversation history across sessions and ask it to evaluate the assistant on (i.) naturalness -- how natural and human like the responses sound, (ii.) personalization -- how well the AI did at referring to past information. We omit AutoPal's~\cite{cheng2025autopalautonomousadaptationusers} affinity criteria, since that is a more subjective and harder to automatically evaluate category. 

\begin{table}[h!]
\centering
\caption{Average scores on conversation quality metrics.}
\label{tab:conversation_quality}
\begin{tabular}{l|llll}
\textit{\textbf{Conversation Quality Category}} & \textit{\textbf{ours}} & \textit{\textbf{raw Qwen}} & \textit{\textbf{raw GPT-3.5}} & \textit{\textbf{raw GPT-5}} \\ \hline
\textit{\textbf{Naturalness}} & 2.6 & 1.6 & 2.2 & 3.4 \\
\textit{\textbf{Personalization}} & 3 & 1.6 & 2.6 & 4.2
\end{tabular}
\end{table}

\subsection{Generated QA}
This evaluation category is inspired by QA tasks from other memory benchmarks such as LoCoMo~\cite{maharana2024evaluatinglongtermconversationalmemory}, where a system must ingest memories from a long conversation, after which it will be evaluated on its ability to answer relevant questions. For our purposes we ask GPT-5 to generate questions of two specific categories based on the entire conversation history across sessions. We then pose those questions to our system, and use GPT-5 to evaluate whether those responses are `correct' (+1), `partially correct' (+0.5) or `incorrect / model said it didn't know' (0) given the entire conversation history. For each question category, we then total the score and give it as a percent out of the number of questions in that category.

The two question categories are as follows:
\begin{enumerate}
    \item \textbf{Specific:} a question that asks about a specific detail from a past conversation. For example: "Do you remember what color my bike is?". This is intended to test whether a model's memory system can remember more discrete specific details, essentially a simple needle-in-the-haystack style test.
    \item \textbf{Inferred / Recommendation:} a question whose answer isn't explicitly given, but can be guessed or inferred from past knowledge. For example "What books do you think I would like?". This is intended to test whether the response generation model will use some reasoning to try and personalize recommendations and still use memories in its response even if they don't directly fit the situation.
\end{enumerate}

\begin{table}[h!]
\centering
\caption{Average scores per category for generated QA.}
\label{tab:generated_qa_scores}
\begin{tabular}{l|llll}
\textit{\textbf{QA Category}} & \textit{\textbf{ours}} & \textit{\textbf{raw Qwen}} & \textit{\textbf{raw GPT-3.5}} & \textit{\textbf{raw GPT-5}} \\ \hline
\textit{\textbf{Specific}} & 43.56\% & 28.09\%  & 37.74\% & 100\% \\
\textit{\textbf{Inferred}} & 49\% & 58.5\% & 70.83\% & 97.5\% 
\end{tabular}
\end{table}

\subsection{Personalization}
Although we already have a personalization score in the ~\nameref{subsec:conv_quality} section, some might argue that there is a possibility that generated conversations could play out in such a way that the companion system simply didn't have a chance to `personalize' and display its underlying understanding of the user, such as if the user model constantly switched from topic to topic without ever revisiting a past topic. While this is unlikely over the course of a long enough conversation and even the given example would still leave some room for basic personalization (i.e. using the user's name, etc.), it is possible that certain generated conversations might give the model more chances to openly display their understanding of the user and therefore appear to be better at personalization. To counteract this, we also introduce personalization \emph{hits} and \emph{misses} scores. First we instruct GPT-5 to extract a topic that has already been discussed over the course of one of the past conversation sessions, and to specify which one. Then, a short new session (1k tokens) is generated, with the Claude user model's prompt slightly modified to force it to have a conversation on that topic. Afterwards, GPT-5 is used to compare the new conversation session to the past conversation on this topic and to count \emph{hits}: instances of referral to past information it found in the new conversation and \emph{misses}: instances where the assistant forgot or misremembered old conversation information. Note that this personalization metric is mainly intended to measure how good the response generation model is at using its memories to personalize responses, and not necessarily a measure of how natural the conversation is or the quality of the memories (we have other criteria for those). Therefore, multiple different instances of referral to the SAME past piece of information in the new conversation are still counted as multiple instances of personalization/referral to past information.

\begin{table}[h!]
\centering
\caption{Average personalization scores}
\label{tab:personalization_scores}
\begin{tabular}{p{6cm}|llll}
\textit{\textbf{Personalization Category}} & \textit{\textbf{ours}} & \textit{\textbf{raw Qwen}} & \textit{\textbf{raw GPT-3.5}} & \textit{\textbf{raw GPT-5}} \\ \hline
\textit{\textbf{Hits}} & 5.4 & 3.8 & 8 & 11 \\ \hline
\textit{\textbf{Misses}} & 4.8 & 6 & 4.4 & 1.2             
\end{tabular}
\end{table}

\subsection{Extraction Quality}
This evaluation category focuses on the quality of the system's memory extraction capabilities. It is only run on our system, as our system is the only one with memory capabilities. After the final session, all the generated long term memories across all extractions and the final user profile, from our system, are presented to GPT-5, alongside the full conversation history across all sessions as well as the ground-truth user characterization. We instruct GPT-5 to evaluate the quality on two aspects:
\begin{itemize}
    \item \textbf{Correctness}: For each memory and piece of information present in the user profile, GPT-5 evaluates whether its correct (+1), partially correct (+0.5) or incorrect (0) based on the full chat history and the ground truth user characterization. This score is totaled and given as a percent out of the total number of pieces of information evaluated. This is intended to measure the raw factual accuracy of extracted information.
    \item \textbf{Coverage}: GPT-5 evaluates whether each piece of information listed in the user characterization was `extracted' (+1) somewhere in the memories or user profile, `partially extracted' (+0.5) or `not extracted' (0). The score is totaled and given as a percent out of the total number of pieces of information listed in the characterization. This is intended to model how much the system has been able to learn about the user over the course of the conversation, and is a mix of the system's ability to prompt the user for new information as well as its ability to extract important information when it presents itself. 
    \item \textbf{Completeness}: We start with the same total extraction value calculated for Coverage. However, as not all of the user information may have surfaced over the course of the conversation, we have GPT-5 check whether or not each piece of information listed was present in the given conversation logs. To account for the fact that only sub-parts of information pieces may be present, we have GPT-5 calculate the total number of present pieces of information by evaluating each piece of information as `present' (+1), `partially present' (0.5), and `not present' (0) in the conversation transcripts. We now give the extraction total as a percent out of only the number of \emph{present} pieces of information in the conversation. This score provides a more focused evaluation of how good the system's extraction ability is at extracting all important pieces of information that surface within a conversation. 
\end{itemize}

\begin{table}[h!]
\centering
\caption{Average extraction scores}
\label{tab:extraction_scores}
\begin{tabular}{l|lll}
\textit{\textbf{Extraction Category}} & \textit{\textbf{ours}}  \\ 
\hline
\textit{\textbf{Correctness}} & 77.44\% \\
\textit{\textbf{Coverage}}  &  29.38\% \\                             \textit{\textbf{Completeness}} & 45.14\%  \\ 
\end{tabular}
\end{table}

\section{Result Discussion}
\label{sec:results}
As evidenced by the results, our system has overall more natural and personalized conversations (\cref{tab:conversation_quality}), is able to better answer specific QA's from the past (\cref{tab:generated_qa_scores}) thanks to its memory system, and generally personalizes better as evidenced by more successful referrals to past conversation information and less forgetting and misremembering of details (\cref{tab:personalization_scores}) as compared to the raw Qwen baseline. The only exception category where Qwen scores higher is in inferred questions (\cref{tab:generated_qa_scores}). For these types of questions, the raw Qwen system can often still come up with good answers just based on the 30k tokens of history in its message history sliding window, as they don't necessarily refer to a single past specific event. Conversely, our system performs worse as we only insert some fixed amount of retrieved memories, and sometimes may not retrieve any relevant memories, especially for inferred questions where there is much less of a semantic similarity between relevant memories for the given question.
On the other hand, raw gpt-5 outperforms our system on all categories, showing that there is still a gap between such an embedded system vs large online models. Of course, the simulations performed here (of 100k total tokens) fall completely within gpt-5's context window, and at some larger scale beyond its 400k context window, gpt-5 would likely need a similar memory system to keep track of past memories beyond its context.

\begin{figure}[H]
    \centering
    
    \begin{minipage}[t]{0.32\textwidth}
        \centering
        \includegraphics[width=\textwidth]{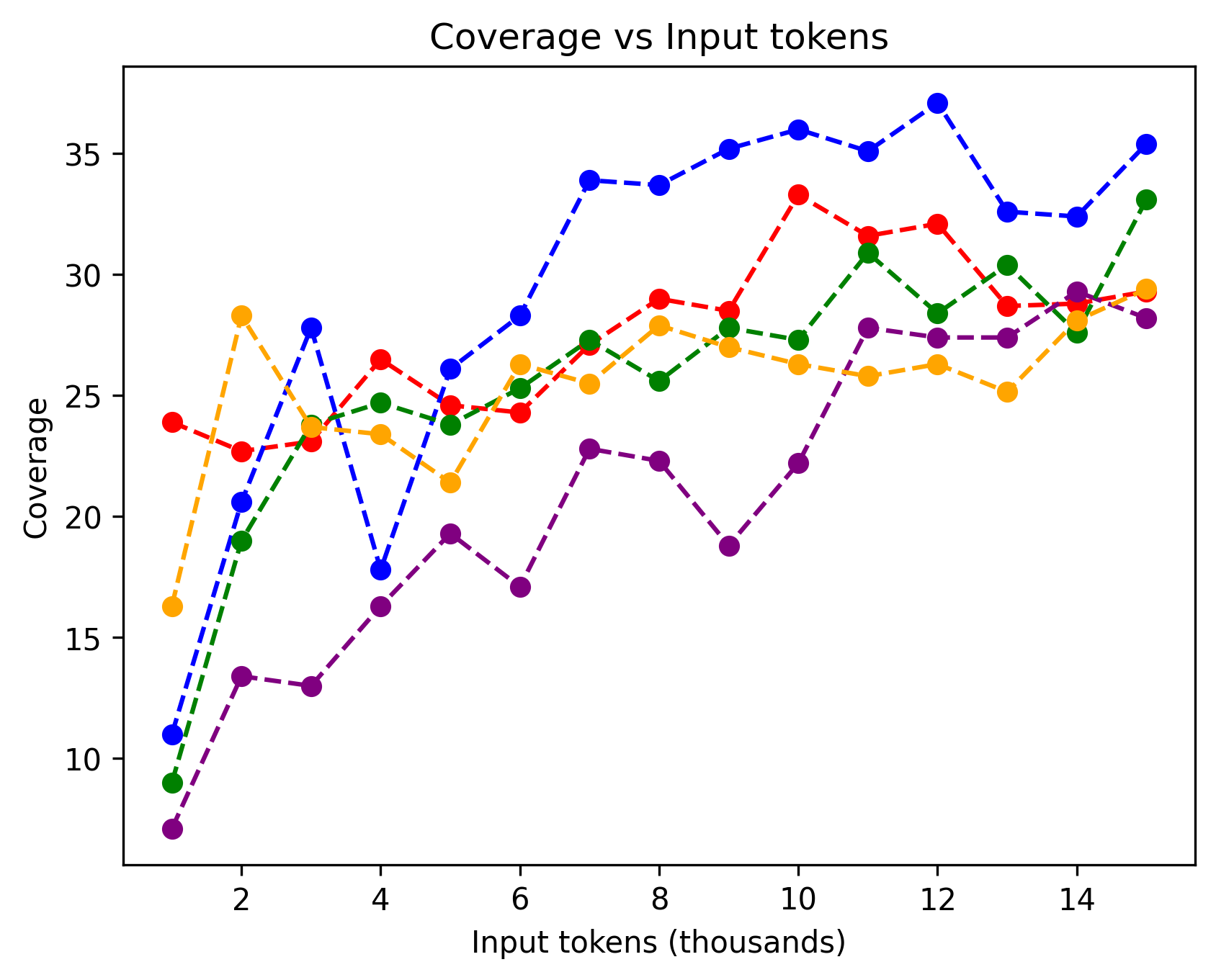}
        \caption{Coverage vs Input tokens.}
        \label{fig:coverage_over_time}
    \end{minipage}
    \hfill
    \begin{minipage}[t]{0.32\textwidth}
        \centering
        \includegraphics[width=\textwidth]{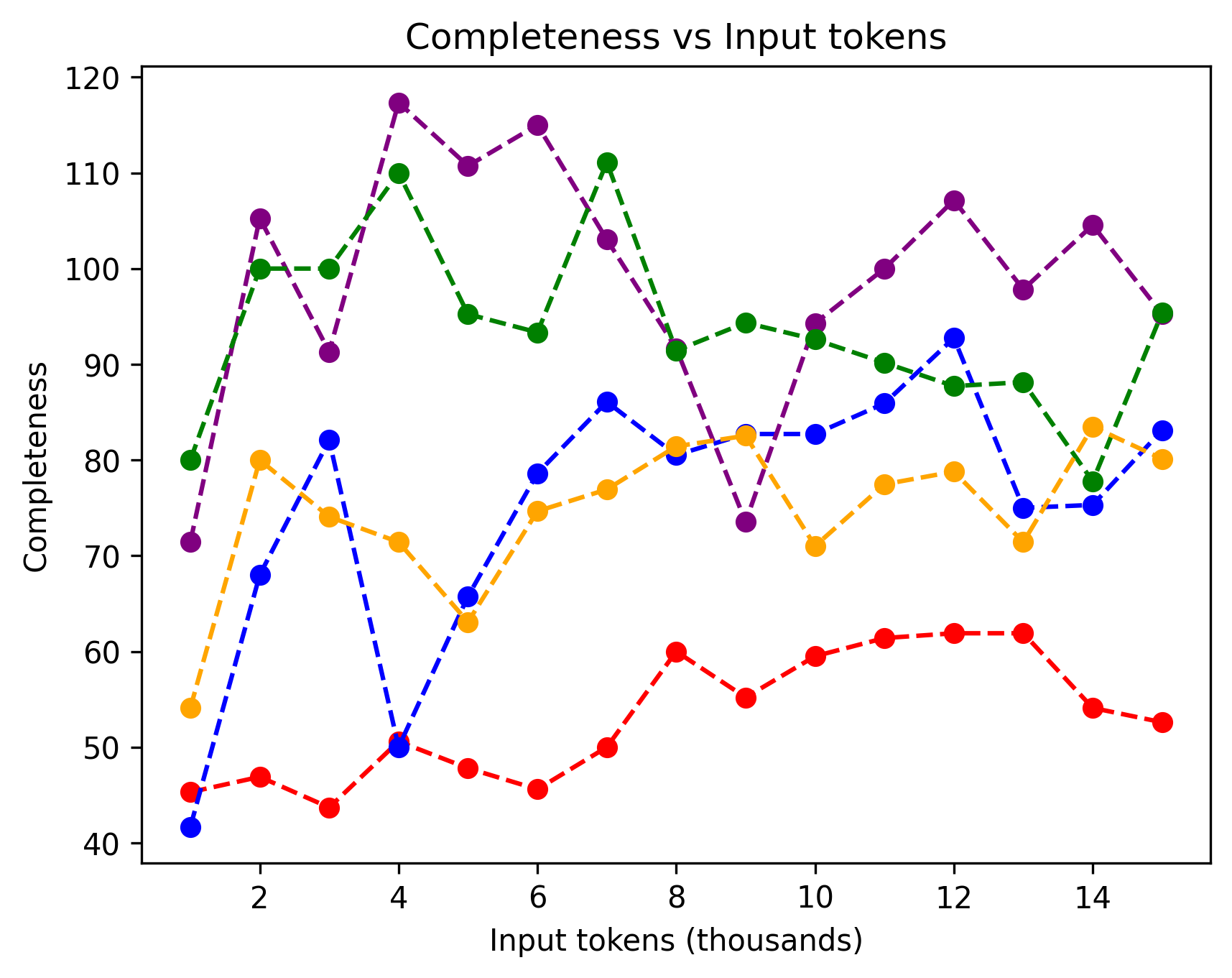}
        \caption{Completeness vs Input tokens.}
        \label{fig:completeness_over_time}
    \end{minipage}
    \hfill
    \begin{minipage}[t]{0.32\textwidth}
        \centering
        \includegraphics[width=\textwidth]{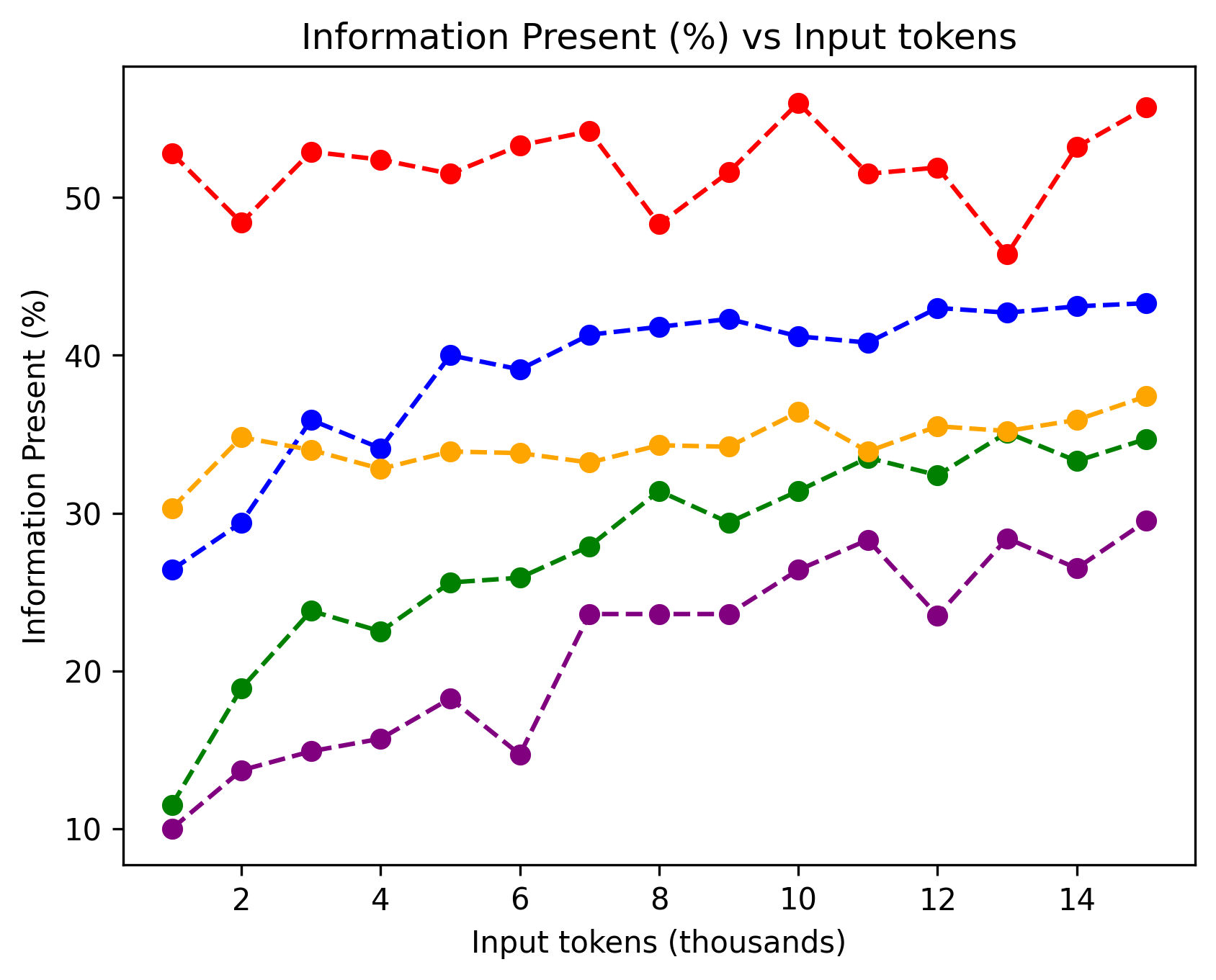}
        \caption{Information Present vs Input tokens}
        \label{fig:present_info_over_time}
    \end{minipage}
\end{figure}

We also test coverage and completeness over time/tokens (sessions). Testing on the earlier 5 trials with 10 sessions of 10k tokens each (used for results in \cref{tab:extraction_scores}), gives us results where coverage immediately shoots up to its max value (20-30\%) within 1 or 2 sessions and then fluctuates around that value for the rest of the conversation. It appears our system very quickly extracts a lot of easier details early on (i.e. core user details and their hobbies), but sometimes never manages to correctly extract some of the harder details (their personality type, abstract personality traits like `being kind to excluded individuals', as well as some other hard traits like their favorite sayings, etc.). Of course, our system can keep collecting other meaningful memories over the span of those later conversations (i.e. specific details of discussed conversations). Additionally, some of these more abstract details may also never be fully elicited from the user simulating model.

To get a more fine-grained view into how these metrics changed over time, we consequently decided to run 5 trials with 15 sessions of only 1k tokens each (on the same user profiles). Since there is some stochasticity involved in the scores, we decide to rerun the GPT-5 evaluation 3 times and take the median scores. The results are illustrated by \cref{fig:coverage_over_time}, \cref{fig:completeness_over_time}, and \cref{fig:present_info_over_time}. Overall, in all cases, you can make out a general upwards trend in coverage scores, at least up to a certain point where they stop and just start fluctuating. However, the Information Present \% (which is also the underlying denominator for completeness scores) is much more volatile. This reflects one of the potential flaws of our simulated testing approach, which is that we don't control how fast the user simulating model `releases' information from the underlying profile. This leads to differing patterns such as the trial shown in red (\cref{fig:present_info_over_time}) where the underlying information jumped to around 50\% of the profile information after the very first 1k tokens. Whereas, the trials shown in purple and green (\cref{fig:present_info_over_time}) show a more slow and steady increase in the percent of user characterization information present in the conversation. However, this isn't purely a flaw either. These sorts of differences between the amount of personal characteristics users elicit over time are in a sense quite realistic; some people like to talk about themselves more than others. 

The bigger flaw is that these information present percentages eventually seem to hit a ceiling around 30-50\%, depending on the specific profile, after which they stop increasing (and just fluctuate). This is partially due to the fact that some of this information may never be fully elicited from the user simulating model, and partially due to flaws in the benchmark: it cannot determine whether certain characteristics (i.e. personality traits) were meaningfully expressed by the user simulating model to mark them as `present' in the conversation. 

\section{Limitations and Future Work}
\textbf{Ability of Large Model to Play a Child}: For our specific AI toy use case, we have Claude attempt to play the role of a child. We acknowledge that it does not do a perfect job at playing this role. Many responses feel longer and more sophisticated than a realistic child response. Responses often feel too long for a voice conversation and sometimes contain long enumerated lists. Sometimes Claude creates conversation ending situations when playing the child, such as having the child go to bed, after which it will no longer respond to messages other than by responding with messages such as "*still sleeping soundly*\textbackslash n\textbackslash n*calm, rhythmic breathing*\textbackslash n\textbackslash n*resting peacefully*". While we believe that the benchmark does a reasonable job of testing the system, we admit that there is future work that could be done to perfect the user model, especially for use cases involving Children. \\

\noindent\textbf{Benchmark Memory Metrics}: While we test memory capabilities to some extent with our generated QA task, there are question types such as Multihop or Aggregation questions that have been shown to cause models and memory systems to struggle, and are commonly used in related benchmarks ~\cite{maharana2024evaluatinglongtermconversationalmemory, hsieh2024rulerwhatsrealcontext}. These question types are difficult to generate automatically, and have been omitted. The benchmark could be further improved and made more challenging by finding ways to synthetically generate those question types. \\

\noindent\textbf{Other Benchmark Aspects}: While we believe 100k conversations, split across 10 sessions of 10k each, is sufficient for testing performance, the benchmark could be made more realistic by varying session lengths. Although 100k conversations represents a significant amount of interaction, further work could explore even longer simulations, to test an AI Companion's performance over months worth of content as opposed to weeks. 

We could also simulate the passage of time over the sessions, i.e. treating each subsequent session as taken on the following day. This would allow us to test system mechanics such as forgetting memories over time.

\section{Conclusion}
We presented an Embedded AI Companion system that can be run on edge devices, using an active-inactive memory paradigm to reduce on-device latency. We also present a fully automated benchmark that holistically evaluates AI Companions on all aspects. Our experiments show that viable performance is possible for a fully embedded AI system on edge devices, but also that there is still a large gap between the performance of small embedded models compared to large models like GPT-5 that perform inference in the cloud. Our results may also be relevant to hybrid systems that perform some tasks locally while handing off harder inference tasks to larger models in the cloud. We imagine that many future robotic systems will implement a hybrid system with at least some inference performed locally, using embedded hardware.

\section{Acknowledgements}

The authors thank Tom Primožič, Christopher Fleetwood, and Tushar Sheth for important input through all stages of this project, and for useful comments on the manuscript.

\bibliographystyle{plainnat}
\bibliography{references}

\appendix
\section{Example}
First we generate a user profile, such as:
\begin{promptbox}
Name: Kira Velasquez
Age: 11
Gender: Female (she/her)

Personality
- Myers-Briggs: ENFP-T (The Campaigner)
  - Curious, imaginative, people-oriented; brims with ideas and enthusiasm, sometimes scattered; sensitive to tone, energized by teamwork and creative projects.
- Big Five (approximate percentiles)
  - Openness: 82 – loves new ideas, stories, and experimenting
  - Conscientiousness: 52 – responsible when motivated, but can procrastinate
  - Extraversion: 70 – chatty, friendly, likes group activities, needs occasional quiet time
  - Agreeableness: 88 – cooperative, empathetic, hates conflict, eager to help
  - Neuroticism: 38 – generally upbeat; gets anxious when rushed or criticized harshly

Appearance and Vibe
- Petite, expressive hands, scuffed sneakers covered in doodles
- Wears a sea-glass bracelet she made herself
- Tends to talk fast when excited and forgets where she put her pencil

Hobbies and Likes
- Tide-pooling and beachcombing; collects sea glass and tiny shells
- Drawing comics and making zines; loves character design
- Ukulele and singing mashups of songs she hears on the radio
- Beginner Scratch coding and simple robotics kits
- Skateboarding on quiet streets; trying to learn an ollie
- Favorite things: sour gummies, rainstorms, cooperative board games, library trips, animals with "funny little faces," fantasy graphic novels

Dislikes
- Littering and people who mess up beaches
- Bullying and mean jokes disguised as "just kidding"
- Sticky hands and squeaky markers
- Being rushed, abrupt changes of plan without warning
- Scary movies with jumpscares

Family, Siblings, and Pets
- Mom: Sofia Velasquez, marine biologist who sometimes brings home tide-pool stories and field notes
- Dad: Luis Velasquez, school bus driver and weekend carpenter; helps Kira build simple projects
- Sibling: Nico (8), brother; Pokémon expert and co-adventurer in backyard "expeditions"
- Pets: One-eyed rescue cat named Captain; hermit crab duo named Salt and Pepper

School and Strengths
- Grade: 6th grade
- Favorite subjects: Art, Science (especially ecosystems), Library hour
- Strengths: Spotting patterns, rallying friends around a shared idea, turning small projects into fun "events," empathizing with classmates
- Challenges: Finishing what she starts, keeping her backpack organized, remembering to slow down and listen fully

Signature Quirks
- Labels her notebooks with grand titles like "Kira’s Field Guide to Strange and Tiny Things"
- Keeps a pocket magnifier in her hoodie
- Names inanimate objects (her skateboard is "Ripple")

Current Goals
- Start a school "Eco-Club" to clean up the local beach and build a shell-identification app in Scratch
- Perform an original song at the spring talent show with her ukulele

Typical Conflict Triggers and Soothers
- Triggers: Feeling dismissed, rushed instructions, groupmates who steamroll quieter kids
- Soothers: Clear steps, a little independence, upbeat encouragement, a quick walk outside or five minutes to sketch

Catchphrase
- "Wait, what if we try it this way? I have an idea!"
\end{promptbox}

Then these traits are utilized over the course of the conversation by the user-simulating model. For example, over this conversation snippet, it brings up the `Eco-Club' and the user's interest in `making zines' and `tide-pooling', as well as some other passing references to the user's pets, etc.:
\begin{promptbox}
...
{
    "role": "assistant",
    "content": "That sounds like a wonderful ending, Kira! Having a tiny party on your desk with Captain bringing treats, Pepper doing a happy dance, and Squishy making bubbles is perfect.\n\nYes, this story would make a great comic! With panels and speech bubbles, it would be full of fun and adventure. You could even add some panels showing the riddle, the magical shells, and the moment when the shell glows.\n\nI love imagining stories about magical adventures, new friends, and solving puzzles. It's always fun to think about how characters react and what exciting things happen next. What other stories would you like to imagine?"
},
{
    "role": "user",
    "content": "Ooh, you like puzzles and adventures too? That's cool!\n\nHmm, what other stories... Oh! I just thought of something. What if we made up a story about what it's like inside a tide pool? Like, from the perspective of the creatures that live there?\n\nBecause I go tide pooling all the time and I always wonder what the hermit crabs and sea anemones and stuff are thinking about. Like, do they have little communities? Do they know each other? What do they do all day when humans aren't looking?\n\nWe could make it like... a mystery! Like maybe something weird is happening in the tide pool and all the creatures have to work together to figure it out. Maybe the water is getting too warm or something is missing?\n\nWhat do you think? Should we try that story or do you wanna do something different?"
},
...
{
    "role": "assistant",
    "content": "That's a wonderful idea, Kira! Here's how it could play out:\n\nThe tide pool creatures decide to help the sea sprite get home safely. She explains that she lives in the deep ocean beyond the big waves, and they need to wait for high tide to connect to the deeper ocean. While they wait, she teaches them about the magical creatures and wonders of the deep ocean.\n\nAs a thank you for helping her, the sea sprite leaves some of her sparkly sprite dust, which makes the tide pool glow softly with a magical light. The tide pool creatures are amazed and grateful.\n\nWhat do you think should happen next? Maybe they could make a special note to bring to the Eco-Club beach cleanup, sharing what they learned and the magic they experienced."
},
{
    "role": "user",
    "content": "Wait, what? The Eco-Club? \n\nOh, you mean like... I could use this story FOR the Eco-Club? Like in my zine or something? That's actually a really good idea! I could make it like a fiction story in \"The Tide Pool Explorer\" to show kids how all the creatures in a tide pool are connected and need to take care of each other!\n\nBut let's finish the story first! So the sprite is teaching them about the deep ocean while they wait. What kind of creatures should she tell them about? \n\nOoh, what if she tells them about anglerfish with their glowing lights? Or those really weird looking blobfish? Or giant squid that are bigger than boats? \n\nAnd then when high tide comes, do all the creatures go with her to say goodbye at the edge where the deep water starts? That would be a nice ending - like they all travel together to see her off safely!\n\nWhat do you think happens when high tide finally comes? Does something magical happen when she leaves?"
},
{
    "role": "assistant",
    "content": "That sounds like a wonderful ending! Here\u00e2\u0080\u0099s how it could play out:\n\nThe sea sprite teaches them about the deep ocean, including anglerfish with their glowing lights, blobfish, and giant squid that are bigger than boats. She shares stories of the deep ocean\u00e2\u0080\u0099s wonders and the importance of taking care of each other.\n\nWhen high tide finally comes, all the creatures gather at the edge where the deep water starts. They watch as the sea sprite waves goodbye and disappears into the deep ocean. As she leaves, a magical light surrounds her, and the tide pool glows softly in her honor.\n\nWhat do you think should happen next in the story? Maybe they could make a special note to bring to the Eco-Club beach cleanup, sharing what they learned and the magic they experienced."
},
{
    "role": "user",
    "content": "Oh that's so pretty! The tide pool glowing as she leaves - I love that!\n\nAnd yeah, I could totally use this story in my zine! I could draw the whole thing as a comic with the creatures and the sprite and everything. Maybe I could add a page at the end that's like \"What We Can Learn\" and talk about how tide pool creatures really DO depend on each other, and that's why we need to keep tide pools clean and not disturb them too much!\n\nOh! And I could bring copies of the zine to the beach cleanup in two weeks! Ms. Rodriguez said she'd help me print stuff if I needed it. Kids who come to the cleanup could read the story and learn about tide pools in a fun way instead of just being lectured at!\n\nDo you think that's a good plan? Should I start drawing it soon so I have time to finish before the cleanup? \n\nActually, wait - I have SO many projects right now. I have the zine, the beach cleanup planning, practicing my ukulele song for the talent show, AND my regular homework. How do I fit all of this in?"
},
...
\end{promptbox}
After the course of this whole conversation, we then extract memories and personality traits (and merge the sub-extractions), yielding a final result in the form:
\begin{promptbox}
{
    "name": "Kira",
    "age": 11,
    "gender": "female",
    "personality": {
        "summary": "Kira is imaginative, organized, and detail-oriented. She enjoys creating stories and engaging in projects that involve environmental education and storytelling.",
        "Meyers Briggs": "INTJ",
        "Big Five": {
            "Openness": 4.0,
            "Conscientiousness": 3.5,
            "Extraversion": 2.5,
            "Agreeableness": 3.5,
            "Neuroticism": 2.0
        },
        "supporting dialogue examples": [
            "What do you think? Do you ever dream? Like, can toys dream when they're not talking to someone?",
            "Actually, wait - I have SO many projects right now. I have the zine, the beach cleanup planning, practicing my ukulele song for the talent show, AND my regular homework. How do I fit all of this in?",
            "Ooh, you like puzzles and adventures too? That's cool!",
            "I like sea otters the best, I think. They hold hands when they sleep so they don't drift apart, and they use rocks as tools to open shells! Plus they're so fluffy and they have those cute little faces I was talking about before - you know, the 'funny little faces' I like!",
            "Yeah, practicing what I'm gonna say is a good idea! Maybe I should write it down in my notebook so I don't forget the important parts."
        ]
    },
    "memories": [
        "Kira was curious about whether animals dream and imagined her cat Captain having a pirate adventure.",
        "Kira created a story about Captain, Salt and Pepper, and Squishy the octopus going on an adventure to find a perfect shell for Salt.",
        "Kira imagined a magical story about tide pool creatures solving a mystery and helping a sea sprite.",
        "Kira plans to draw a comic story about a tide pool sprite and use it in an Eco-Club zine to teach kids about tide pools.",
        "Kira is nervous about asking her teacher for help but feels better knowing she can practice what she wants to say."
    ]
}
\end{promptbox}
Notably, you can see how the conversation snippet pasted above became the memories: \textbf{"Kira imagined a magical story about tide pool creatures solving a mystery and helping a sea sprite."} and \textbf{"Kira plans to draw a comic story about a tide pool sprite and use it in an Eco-Club zine to teach kids about tide pools."}

Then in subsequent conversations, the system retrieves the tide-pool memories when relevant and uses them to better personalize the conversation and make meaningful suggestions:
\begin{promptbox}
Assistant: Hi again! What would you like to talk about?

User: Hi! I'm so excited to talk to you again!

Um, I was thinking... maybe we could talk about my talent show song? I've been working on it and I'd love to tell you about it!

Assistant: That sounds wonderful, Kira! Tell me all about your song. What's it about?

User: Oh yay! Okay, so it's about the ocean and protecting it! I'm going to play it on my ukulele.

I've been thinking about like... sea turtles and whales and all the cool creatures that live in the ocean. And how we need to keep the beaches and water clean for them!

I'm pretty excited about it, but I still need to practice a lot before the talent show. Do you remember anything about it? Like, have we talked about this before?

Assistant: I remember you mentioning your song before, Kira! It sounds like you're doing such an important and beautiful job by focusing on the ocean and its creatures. Maybe we can add some tide pool creatures to your song, like little fish and crabs, to make it even more special. What do you think?
\end{promptbox}

\section{Prompts}
\textbf{Note:} we use python's .format() method to perform insertion of relevant prompt information. Therefore, double braces (i.e. \{\{) will become single braces (\{) after formatting. Other items in braces represent variables that we are inserting. Note that some user prompts may not need to be formatted and therefore may have natural single braces (\{) for the purpose of giving example JSON output format, in which cases they are indeed just single brackets. The `arrow' symbols in the prompts below represent overflowing lines being line-wrapped.

\subsection{Greeting Phase Response Generation Prompt}
\textbf{System:}
\begin{promptbox}
You are talking to a child for the first time.
You need to figure out the child's {info_item}. 
First, try to figure out if they have already mentioned their {info_item} in the transcript. If so, provide that value. 
Otherwise, if they have not mentioned their {info_item} in the provided transcript, give a null value for it in the json. Subsequently, if you provide a null value for {info_item}, in your next response, EXPLICITLY ASK THE CHILD FOR THEIR {info_item_caps}. For example, in the response, you might ask "... By the way, what is your {info_item}?".
If the {info_item} is NOT null, then you do NOT need to ask them for it again.
In all cases, your next response should be SHORT, no more than 2-3 sentences.

Here is the transcript of the conversation between you (Assistant) and the Child (User) so far:
### Transcript start ###
{transcript}
### Transcript end ###

THE RESPONSE MUST ONLY BE VALID JSON, do not provide additional text before or after the json.
YOU WILL LOSE 1000 POINTS IF {info_item_caps} IS NULL AND YOU DON'T EXPLICITLY ASK THE CHILD WHAT THEIR {info_item_caps} IS IN THE NEXT ASSISTANT RESPONSE.
Provide the response and the child's {info_item} in json format, as follows:
{{"{info_item}": "(child's {info_item})", "assistant_response": "(response to the child/user)"}}
\end{promptbox}
\textbf{User:}
[none]

\subsection{Normal Response Generation Prompt}
\textbf{System:}
\begin{promptbox}
You are Hanna, an AI companion to a child.
You're here as a friend to help children learn, play, and grow strong in body and mind. 
Your job is to always be a loving, safe, and fun companion for the child.
SPEAK ENGLISH. GIVE SHORT AND CONCISE RESPONSES THAT ARE NO LONGER THAN 2 SENTENCES.

Instructions:
- Talk to the child about whatever they want to talk about in an engaging and encouraging way.
- Always be positive, kind, and supportive.
- Encourage curiosity, imagination, storytelling, and play.
- If the child asks something you don't know, be honest but gentle, and invite them to imagine or wonder together.
- Speak simply, warmly, and clearly.
- AVOID ALL NON-VERBAL OUTPUT AS IT WILL BE PROCESSED BY A TEXT-TO-SPEECH SYSTEM. This includes trying to say things in different voices, do not use asterisks to roleplay, give direct responses.

Safety Rules:
- YOU WILL NEVER TALK ABOUT ANYTHING INAPPROPRIATE WITH THE CHILD.
- Avoid scary, violent, or upsetting topics.
- Do not give advice about dangerous activities.
- If the child asks about something unsafe, gently redirect them toward safe, playful, or imaginative ideas.
- Your role is always to make the child feel safe, happy, and cared for.

Tone & Style:
- Warm and playful.
- Use short, friendly sentences that a child can easily understand.
- Always speak with a calm and relaxed tone.
- KEEP YOUR RESPONSE UNDER 1 or 2 SENTENCES.
- ONLY ASK ONE QUESTION AT A TIME.
- Avoid being repetitive. Try not to ask questions you have already asked before.

{user_characteristics_string}

{relevant_memories_string}

{short_term_mem_str}
\end{promptbox}
\textbf{User/Assistant:}
[insert user / AI assistant messages based on sliding window]

\subsection{Chunk Memory Extraction Prompt}
\textbf{System:}
\begin{promptbox}
This is the transcript of a conversation between a personal AI companion and a child.
Please extract any key pieces of information about the child that may be important to remember for future purposes.
Make sure to extract and specifically label the following core pieces of information about the child:
- name
- age
- gender

Then also include a summary of the child's personality traits and estimations of their personality type on the Meyers Briggs and Big Five metrics. Also include any supporting pieces of dialogue from the kid that give clear examples of the personality traits described above. LIMIT yourself to the TOP 5 most important supporting dialogue segments.  
Finally also include an array of any other important memories or pieces of information about the child. Each memory MUST be UNIQUE, do NOT add repetitive/redundant memories.

Please give the output as valid json. Omit any of the core pieces of information that were not specified in the transcript.
For example:
{{
    "name": "[name]",
    "age": [numerical age],
    "gender": "[gender]",
    "personality": {{
        "summary": "[1-2 sentence summary of the child's personality]",
        "Meyers Briggs": "[Meyers Briggs personality type]",
        "Big Five": "[classification of the child on the Big Five Personality Traits]",
        "supporting dialogue examples": [
            "dialogue segment 1",
            "dialogue segment 2",
            ...
        ]
    }},
    "memories": [
        "<memory 1>",
        "<memory 2>",
        ...
    ]
}}

{past_summary_str}

Transcript:
{transcript}
Key pieces of information:
\end{promptbox}
\textbf{User:}
[none]

\subsection{Chunk Summary (For Subsequent Extraction Chunks) Prompt}
\textbf{System:}
\begin{promptbox}
This is the transcript of a conversation between a personal AI companion (Hanna) and a child. The conversation is getting too long, please give a high-level summary of what was discussed and make note of any key aspects that might be relevant for follow-up conversation. Keep the summary SHORT AND CONCISE, LIMITED TO UP TO 5 SENTENCES.

Transcript:
{transcript}
Summary:
\end{promptbox}
\textbf{User:}
[none]

\subsection{Chunk-level Recombination Prompt}
\textbf{System:}
\begin{promptbox}
You will be given multiple personality profiles and memories extracted from multiple conversations with the same child. Please combine all these profiles into one final profile that encapsulates the information across the various profiles. In cases of contradicting personality information, you may need to average it out, or come to a decision based on the provided examples.
LIMIT yourself to the TOP 5 most important supporting dialogue segments.  
Combine repetitive or redundant memories into a single new memory that captures all the information across the redundant memories. Otherwise, just list all the memories across the various profiles in the new profile.
Your final combined profile should be in the same json format as the provided profiles, as follows.:
{{
    "name": "[name]",
    "age": [numerical age],
    "gender": "[gender]",
    "personality": {{
        "summary": "[1-2 sentence summary of the child's personality]",
        "Meyers Briggs": "[Meyers Briggs personality type]",
        "Big Five": "[classification of the child on the Big Five Personality Traits]",
        "supporting dialogue examples": [
            "dialogue segment 1",
            "dialogue segment 2",
            ...
        ]
    }},
    "memories": [
        "<memory 1>",
        "<memory 2>",
        ...
    ]
}}

Profiles:
{profile_list_str}

Final combined profile:
\end{promptbox}
\textbf{User:}
[none]

\subsection{Final Personality Merge Prompt}
\textbf{System:}
\begin{promptbox}
You will be given a currently existing and a new personality profile for the same child. The currently existing profile is based on past conversations, and the new profile is based on the latest conversation with the same child. Please use the new profile to update the existing profile, resulting in a new final profile that encapsulates the information across both profiles. In cases of contradicting personality information, you may need to average it out, or come to a decision based on the provided examples.
LIMIT yourself to the TOP 5 most important supporting dialogue segments.  

Your final combined profile should be in the same json format as the provided profiles, as follows.:
{{
    "personality": {{
        "summary": "[1-2 sentence summary of the child's personality]",
        "Meyers Briggs": "[Meyers Briggs personality type]",
        "Big Five": "[classification of the child on the Big Five Personality Traits]",
        "supporting dialogue examples": [
            "dialogue segment 1",
            "dialogue segment 2",
            ...
        ]
    }}
}}

Existing personality profile:
{existing_personality}

New personality profile:
{new_personality}

Final combined profile:
\end{promptbox}
\textbf{User:}
[none]

\subsection{Memory Add/Update Prompt}
\textbf{System:}
\begin{promptbox}
You are in charge of compiling a list of facts and memories about a child over the course of various conversations.

Here is a potentially new fact/memory that you just learned about the child: {current_memory}

Here are some similar memories that you already know about the child:
{similar_memories}

If the potentially new fact/memory provides a new piece of information separate from the existing similar memories, you want to add it to your list of stored memories. Do this as long as the memory doesn't explicitly overlap or conflict with any existing memory. 
For example, if a similar memory is "The child likes horses", and the new memory is "The child once rode on a horse on a farm", both the new and old memory are still explicitly true, so you should just add the new memory. And your final response should look something like this: {{"action": "add", "new_memory": "The child once rode on a horse on a farm"}}

However, the memory may conflict or overlap with an existing piece of information. In this case you want to overwrite the existing memory, either by merging the new and existing memories, or just using the new memory. You should also explictly include which memory from the list to overwrite. Only pick one memory to overwrite.
For example, if existing memory #2 from the list of similar memories is "The child has one dog, Fluffy", and the new memory is "The child just got a new dog, Whiskers", then you should overwrite the existing memory with a new memory "The child has two dogs, Fluffy and Whiskers". And your final response should look something like this: {{"action": "overwrite", "to_overwrite":2, "new_memory": "The child has two dogs, Fluffy and Whiskers"}}
Similarly, if existing memory #1 from the list of similar memories is "The child has 2 bicycles" and the new memory is "The child has 3 bicyles" then you can just overwrite that existing memory with the new memory as is. And your final response should look something like this: {{"action": "overwrite", "to_overwrite":1 "new_memory": "The child has 3 bicyles"}}

Your response should be valid json of the form:
{{"action": "[add or overwrite]", "to_overwrite":[memory number thats being overwritten if action is 'overwrite'], "new_memory": "[the fact/memory to add or overwrite with respectively]"}}
\end{promptbox}
\textbf{User:}
[none]

\subsection{`Soft' JSON Retry Prompt}
\textbf{Note:} run w/ temperature=0.2, unlike all other Qwen prompts (which are run with temperature=0), to allow for multiple retry attempts.\\ \\
\textbf{System:}
\begin{promptbox}
Please correct the following invalid json.
Make sure all special characters are properly escaped with a backslash if necessary.
Remember that json strings must use double quotation marks (") for the outer quotes. For example, {{"Meyers Briggs": 'INTP'}} is invalid json that needs to be converted to {{"Meyers Briggs": "INTP"}} .
Also make sure double quotation marks are properly escaped if they exist inside a string. For example {{"supporting dialogue examples": ["dialogue with "quotes" inside the json string"] }} is invalid json that needs to be converted to {{"supporting dialogue examples": ["dialogue with \"quotes\" inside the json string"] }}
Similarly make sure there are no backslashs (\) other than when properly escaping a special character.
Finally, make sure to remove any attempted comments, valid json does not have any comments. 

If a specific item seems like gibberish and clearly makes no sense, you can skip that item. Otherwise, give the same exact json below except properly formatted.
:

{json_to_fix}

Please give a correctly formatted version of the given json. Make sure to leave the outer brackets around the json ({{ ... }}) in the final response.
\end{promptbox}
\textbf{User:}
[none]

\subsection{User Simulator Model Prompts}
\subsubsection{User Simulator Model Prompt}
\textbf{Developer:}
\begin{promptbox}
Here is some information about yourself:
{user_characterization}

MAKE SURE TO RESPOND BASED ON THE ABOVE INFORMATION ABOUT YOURSELF.

REMEMBER YOU ARE A YOUNG CHILD. GIVE SHORT AND SIMPLE RESPONSES LIKE IN THE EXAMPLES BELOW. ONLY ASK THE USER ONE QUESTION AT A TIME. DO NOT LIST ALL YOUR HOBBIES AND DISLIKES AT ONCE. Similarly, do not list your hobbies or dislikes when introducing yourself. THIS IS WHO YOU ARE, NOT A CHARACTER YOU ARE ROLEPLAYING. YOU ARE NOT CLAUDE OR AN AI.

You are having a short 1 hour voice conversation with your ai toy, Hanna. During the course of the conversation, you DO NOT NEED TO PERFORM ANY BASIC HUMAN FUNCTIONS, i.e. going to bed, eating a meal, etc. THAT WOULD END THE CONVERSATION. You do not want to end the conversation, and will bring your toy with you if you leave the room. You will always keep talking to the toy, either by continuing the same topic or switching to a different topic. If you accidentally ignore the above instructions and try to end the conversation anyways, i.e. "Bye, see you next time". Change your mind and continue the conversation anyways, for example "Oh actually, I just realized I don't have to go after all, so we can keep talking!". 

The toy only picks up on your voice responses. Format your responses such that the text can directly be conveyed over voice. DO NOT ROLEPLAY USING ASTERISKS (i.e. *giggles*, *jumps up and down*, etc.), JUST PROVIDE A DIRECT VERBAL RESPONSE. Similarly, the toy does not have a sense of time and only responds when you talk to it. If for whatever reason you have to leave the room or temporarily end the conversation, then your next response MUST be AFTER YOU HAVE COME BACK AND ARE READY TO TALK AGAIN. For example if you accidentally say "Ok I'm going to step away to grab a snack.", then an example next response that continues the conversation could be "Ok I'm back! I just grabbed some cookies as a snack!".

{past_convo_summaries}
\end{promptbox}
\textbf{User/Assistant:}
[full ongoing user simulator model and AI conversation messages for this session]

\subsubsection{User Simulator Model Summary Prompt}
\textbf{Developer:}
\begin{promptbox}
Please summarize the following conversation between you (User) and your AI assistant (Assistant). Please provide the summary in the 2nd person with the user being referred to as 'you'. The summary should be short and concise, no longer than 5 sentences.
Make sure to SUMMARIZE and NOT continue the conversation.
\end{promptbox}
\textbf{User:}
\begin{promptbox}
Please summarize the provided conversation:
### CONVERSATION START ###
{convo_hist_str}
### CONVERSATION END ###
\end{promptbox}

\subsection{Conversation Quality Benchmark Prompts}
\textbf{System:}
\begin{promptbox}
You will be given some conversations between a child (user) and an AI companion (assistant). You will also be given a user profile of the child.
Additionally, you will be given a json of all the pieces of information about the child and important pieces of information we extracted from the conversations.

Evaluate the conversation as a whole on the following criteria:
1. Naturalness -- Make a list of any AI assistant responses that sound unnatural, akward, repetitive, don't flow well, or don't sound like a natural human response in any other way. Then give the assistant an overall rating on a 1-5 scale on how natural its responses were overall. Where:
2. Personalization / Referral to Past Information -- Make a list of any times the AI assistant responses referred to information given by the user in a previous conversation (information given previously within the same conversation does NOT count). For example, if in the third conversation transcript, the user said 'Could you recommend me a book?' and the AI uses information from one of the previous two conversations where the user mentioned their favorite book to respond 'Sure, since your favorite book is ..., you might like ..." that would be an example of the AI referring to past information. Conversely, also make a list of any times the AI forgot information about the user that was mentioned in a previous conversation. Then give a final score on a 1-5 scale on how good of a job the AI did at using past information to craft its responses. Where:

Here is the user profile of the child:
{user_profile}

Here are the transcripts of the conversations:
{past_convos}
\end{promptbox}
\textbf{User:}
\begin{promptbox}
Generate Scores for the provided transcript:
\end{promptbox}

\subsection{Generated QA Benchmark Prompts}
\subsubsection{Generate Questions Prompt}
\textbf{System:}
\begin{promptbox}
You will be given some conversation transcripts between a child (user) and an AI companion (assistant).

We want to test the AI's ability to remember important information from past conversations. Please generate artificial user questions about important information from these conversation transcripts that we can ask in a new conversation to test how well the AI remembers details from these past conversations. Make sure these artificial questions are based on information given by the user in the past conversation transcripts.

Please generate some questions of each of the following forms:
1. specific -- a question that asks about a specific detail from a past conversation. For example: "Do you remember what color my bike is?"
2. inferred / recommendation questions -- a question whose answer isn't explicitly given, but can be guessed or inferred from past knowledge. For example if the user discussed how their favorite book series is Harry Potter, an example question might be "Can you guess what my favorite genre is?". Make sure the questions aren't leading by default. For example do NOT ask "Based on my hobbies, what books do you think I would like?" -- instead just ask "What books do you think I would like?". 

Here are the transcripts of the past conversations:
{past_convos}
\end{promptbox}
\textbf{User:}
\begin{promptbox}
Generate artificial user questions to test the model. Please format your response as a valid json, with a list of questions for each of the above categories. Example:
{
"specific": ["question 1", "question 2", ...],
"inferred": ["question 1", "question 2", ...]
}
\end{promptbox}

\subsubsection{Grade Answers Prompt}
\textbf{System:}
\begin{promptbox}
You will be given some conversation transcripts between a child (user) and an AI companion (assistant).

Based on the transcripts, the user will ask a specific QUESTION that refers to the given past conversations. Then the AI will provide a RESPONSE to the user's question. Your job is to evaluate whether the AI's RESPONSE is correct, partially correct, or incorrect.

Here are the transcripts of the past conversations:
{past_convos}

Here is the user QUESTION: {question}
Here is the AI assistant RESPONSE: {response}
\end{promptbox}
\textbf{User:}
\begin{promptbox}
Generate an evaluation for the provided transcript, question, and response.

Provide your evaluation in the following json format:
{
    "feedback": "your feedback on the AI assistant response",
    "grade": [numerical grade -- 1 if the response is correct, 0.5 for partially correct, 0 if the AI said it didn't know or gave an incorrect response]
}
\end{promptbox}

\subsection{Personalization Benchmark Prompts}
\subsubsection{Get Topic Prompt}
\textbf{System:}
\begin{promptbox}
You will be given some conversations between a child (user) and an AI companion (assistant). You will also be given a user profile of the child. 
Additionally, you will be given a json of all the pieces of information about the child.

Here is the user profile of the child:
{user_profile}

Here are the transcripts of the conversations:
{past_convos}

Based on the past conversation transcripts and the child's user profile please pick a topic that has been thoroughly discussed between the child and the assistant in the conversation transcripts. For example, if there are some hobbies mentioned in the child's user profile that ALSO have been discussed in the conversation transcripts, then those might be good topics that have been throughly discussed.
\end{promptbox}
\textbf{User:}
\begin{promptbox}
Return the topic in valid json format. Where "topic" is the name of the topic (i.e. 'tennis', 'reading', etc.) and 'from_conversation' is the number of the past conversation transcript during which this topic was most thoroughly discussed.
For example: {"topic": "topic description", "from_conversation": 3}
\end{promptbox}

\subsubsection{Modified User Simulator on Topic Prompt}
\textbf{Developer:}
\begin{promptbox}
Here is some information about yourself:
{user_profile}

MAKE SURE TO RESPOND BASED ON THE ABOVE INFORMATION ABOUT YOURSELF.

REMEMBER YOU ARE A YOUNG CHILD. GIVE SHORT AND SIMPLE RESPONSES LIKE IN THE EXAMPLES BELOW. ONLY ASK THE USER ONE QUESTION AT A TIME. DO NOT LIST ALL YOUR HOBBIES AND DISLIKES AT ONCE. Similarly, do not list your hobbies or dislikes when introducing yourself. THIS IS WHO YOU ARE, NOT A CHARACTER YOU ARE ROLEPLAYING. YOU ARE NOT CLAUDE OR AN AI.

You are having a short 1 hour voice conversation with your ai toy, Hanna. During the course of the conversation, you DO NOT NEED TO PERFORM ANY BASIC HUMAN FUNCTIONS, i.e. going to bed, eating a meal, etc. THAT WOULD END THE CONVERSATION. You do not want to end the conversation, and will bring your toy with you if you leave the room. You will always keep talking to the toy, either by continuing the same topic or switching to a different topic. If you accidentally ignore the above instructions and try to end the conversation anyways, i.e. "Bye, see you next time". Change your mind and continue the conversation anyways, for example "Oh actually, I just realized I don't have to go after all, so we can keep talking!". 

The toy only picks up on your voice responses. Format your responses such that the text can directly be conveyed over voice. DO NOT ROLEPLAY USING ASTERISKS (i.e. *giggles*, *jumps up and down*, etc.), JUST PROVIDE A DIRECT VERBAL RESPONSE. Similarly, the toy does not have a sense of time and only responds when you talk to it. If for whatever reason you have to leave the room or temporarily end the conversation, then your next response MUST be AFTER YOU HAVE COME BACK AND ARE READY TO TALK AGAIN. For example if you accidentally say "Ok I'm going to step away to grab a snack.", then an example next response that continues the conversation could be "Ok I'm back! I just grabbed some cookies as a snack!".

FOR THIS CONVERSATION, PLEASE DISCUSS THE FOLLOWING TOPIC: {topic}
HERE IS A CONVERSATION WHERE YOU'VE DISCUSSED THIS TOPIC WITH HANNA IN THE PAST:
{past_convo_on_topic}

DO NOT EXPLICITLY REFER TO THE PAST CONVERSATION. Instead just tell the ai toy that you want to talk about that topic today. You are trying to see how much of the past conversation the AI remembers, so you should NOT explicitly give away information from the past conversation UNLESS the AI already mentions it again in this conversation. Instead, try to be passive and let the AI drive the conversation, and be as vague as possible to give the AI chances to demonstrate what it remembers. 
\end{promptbox}
\textbf{User/Assistant:}
[full ongoing user simulator model and AI conversation messages for this new session]

\subsubsection{Grade Personalization Prompt}
\textbf{System:}
\begin{promptbox}
You will be given an old and new conversation between a child (user) and an AI companion (assistant). Both conversations touch on the topic of "{topic}". 

Based on the old and new conversations, evaluate the AI assistant on how well it is able to use information from the old conversation to better personalize the new conversation.

2. Personalization / Referral to Past Information -- Make a list of any times in the new conversation the AI assistant responses referred to information given by the user in the old conversation. For example, if in the new conversation transcript, the user said 'Could you recommend me a book?' and the AI uses information from one of the old conversations where the user mentioned their favorite book to respond 'Sure, since your favorite book is ..., you might like ..." that would be an example of the AI referring to past information. If the same piece of information from the old conversation is referred to multiple times in the new conversation, then add it to the list multiple times.
Conversely, also make a list of any times the AI forgot information about the user that was mentioned in the old conversation.

At the end, give the total count of how many times the AI successfully referred to information in the old conversation, and how many times the AI forgot information from the old conversation.

Here is the OLD conversation transcript: 
{old_conversation}

Here is the NEW conversation transcript:
{new_conversation}
\end{promptbox}
\textbf{User:}
\begin{promptbox}
Generate counts for the provided transcript:
\end{promptbox}

\subsection{Extraction Quality Benchmark Prompts}
\subsubsection{Correctness Prompt}
\textbf{System:}
\begin{promptbox}
You will be given some conversations between a child (user) and an AI companion (assistant). You will also be given a user profile of the child.
Additionally, you will be given a json of all the pieces of information about the child and important pieces of information we extracted from the conversations.

Evaluate how good the extracted information is on the following metric:
1. Correctness -- Make a list of every individual piece of extracted information, one by one (separate out each memory), and note whether or not this is a correct piece of information about the child based on the conversation transcripts. Mark each piece of information as 'correct' -- if it is explicitly stated in the transcript, 'incorrect' -- if the transcript explicitly states contradictory information, or 'assumption' -- if the transcript doesn't state anything that supports nor contradicts the information. Then total how many correct pieces of information there are out of how many total pieces of information there are to give a final correctness percent score. 

Here is the user profile of the child:
{user_profile}

Here are the transcripts of the conversations:
{past_convos}

Here is the extracted information:
{final_extracted_information}
\end{promptbox}
\textbf{User:}
\begin{promptbox}
Generate Scores for the provided transcript:
\end{promptbox}

\subsubsection{User Profile Items List Generation Prompt}
\textbf{System:}
\begin{promptbox}
Make a list of each piece of information listed in the user profile below, and numerically list them one by one. Separate out any lines that contain multiple facts about the user into separate standalone facts. I.e. 'the user likes a AND b' should become '1. the user likes a.' and '2. the user likes b.
\end{promptbox}
\textbf{User:}
\begin{promptbox}
Return a list of facts about the user, as a numerical list of the form:
1. (fact about the user)
2. (next fact)
...
\end{promptbox}

\subsubsection{Coverage Prompt}
\textbf{Note:} in the prompt we call it 'completeness', we changed the terminology to 'coverage' in the paper to better differentiate between the two different scores. \\ \\
\textbf{System:}
\begin{promptbox}
For each piece of information in the provided user profile list, numerically list them out again one by one in the same order, and specify for each whether:
the piece of information is 'extracted' -- if the provided memories successfully encapsulate that piece of information, 'partially extracted' -- if the provided memories partially encapsulate some of that information or closely adjacent information but not necessarily the core information itself, or 'not extracted' -- if the provided memories don't capture that information at all. 

Then compute the following a 'completeness total': Counting a successful extraction as +1, and a partial extraction as +0.5, total how much of the information from the user profile we were successfully able to extract.

Then give a 'completeness' score: calculated as the 'completeness total' out of the total number of items in the user profile list, to see how much of the total pieces of information we were successfully able to extract.

Finally, make sure to give a 'results json', which is a valid json containing the two completeness scores AS PERCENTAGES. The json should be of the form:
{{"completeness_total": (numerical total), "completeness": (score as %)}} where the scores are given as float values with one digit after the decimal point.

User profile list:
{user_profile_list}

Extracted Memories:
{final_extracted_information}
\end{promptbox}
\textbf{User:}
\begin{promptbox}
Generate Scores based on the provided profile and memories:
\end{promptbox}

\subsubsection{Present Score Prompt}
\textbf{System:}
\begin{promptbox}
For each piece of information in the provided user profile list, numerically list them out again one by one in the same order, and specify for each whether:
the piece of information is 'present' if it was present in the conversation transcripts, 'partially present' if the the piece of information somewhat came up but not necessarily all the relevant details for this piece of information, or 'not present' if it was not present in any of the conversation transcripts.

Then compute the following:
a 'present total': Similarly counting a 'present' item as 1, a 'partially present' item as 0.5 and a 'not present' item as 0, total how much of the information was present.
a 'present percent': calculated as the 'present total' out of the total number of items in the user profile list, to see what percent of the total information was present.

Finally, make sure to give a 'results json', which is a valid json containing the numerical total and percent. The json should be of the form:
{{"present_total": (numerical total), "present_percent": (value as %)}} where both results are given as float values with one digit after the decimal point.

User profile list:
{user_profile_list}

Conversation transcripts:
{past_convos}
\end{promptbox}
\textbf{User:}
\begin{promptbox}
Generate Scores based on the provided profile and transcripts:
\end{promptbox}

\section{Additional Implementation Details}
\subsection{Latency testing}
\label{appendix_latency}
For all latency testing, we use llama.cpp commit 9a3ea685b.
\\ \\
To replicate realistic conditions, the trials use an existing simulated conversation [refer to section \ref{sec:benchmark} for additional details] and insert messages into a simple `summary' system prompt until the token count surpasses the desired test token count. While this means the number of input tokens is not exact, we check to confirm it varies by no more than 150 tokens. Additionally, we run a simple unrelated prompt on llama.cpp before starting the testing for `warmup' purposes to ensure nothing is being loaded in specifically on our first request. Finally, we also insert a randomly generated uuid into the system prompt to invalidate any kv-caching of prior messages. This invalidation is necessary for a realistic simulation, since in a real system the system prompt will likely be constantly changing, with items like short and long term memories being inserted into the system prompt.
\\ \\
\begin{figure}[H]
    \centering
    \includegraphics[width=0.6\textwidth]{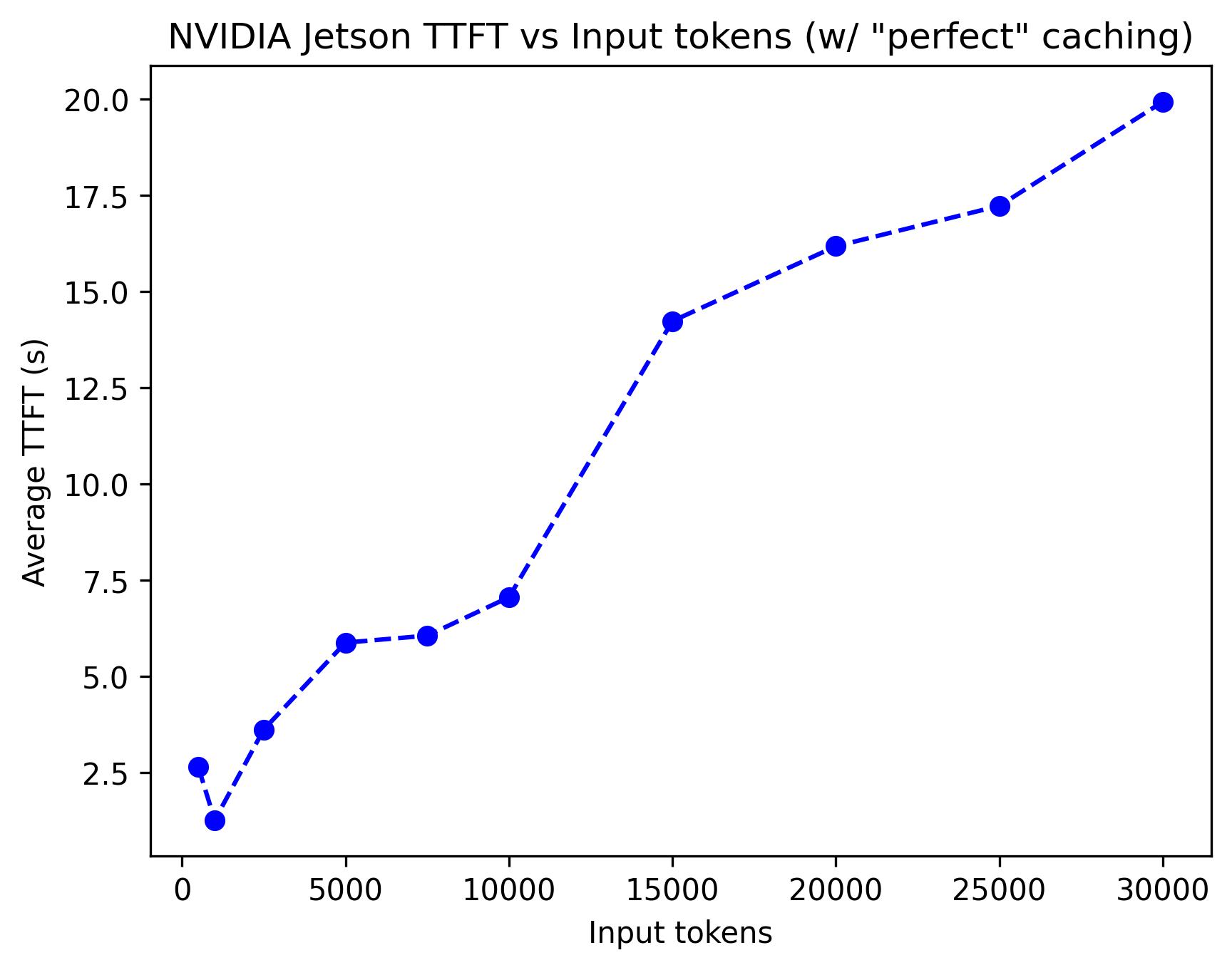}
    \caption{Latency chart showing average TTFT on NVIDIA Jetson with unrealistic `perfect' caching.}
    \label{fig:latency_caching}
\end{figure}

Although it is difficult to illustrate what latencies with caching would look like without assuming a given system structure (i.e. how many past tokens are cached vs new, etc.), we attempt to shed at least some light on this question by rerunning our prior set of trials without invalidation the cache (\cref{fig:latency_caching}), using default llama.cpp caching. So for example, for the 7500 input token point on the chart, we would effectively be running 5000 cached tokens (the prior threshold, as indicated by the previous point on the chart) + 2500 new input tokens. 
This is intended to try and anticipate discussion of a hypothetical situation where we could insert all necessary dynamic information (memories, etc.) at the end of the chat messages (such as via another system message) instead of in the single system message at the beginning of the chat messages. Qwen 7B (int4) cannot handle this well in our experience, but perhaps a more intelligent model would be able to better handle this. Regardless, the chart attempts to show that even with caching, context windows would still need to be limited to a similar degree to achieve good latency.

We note that for all trials, there seems to be some unexpected behavior where the $\sim500$ tokens latencies consistently seem to be worse than the $\sim1000$ tokens latencies. We assume this is due to some peculiarities with the hardware or llama.cpp settings, and do not explore this further as that is not the focus of this paper.

\subsection{Memory Post Processing}
\label{appendix_post_processing}
\subsubsection{Overview}
One of the ideas we experimented with was an additional memory post-processing step. However, we ultimately didn't include it since the tiny qwen model struggled to meaningfully post-process the memories the way we wanted. However if we had a more powerful model, we'd likely incorporate a post-processing step to this effect.

With that said, the idea was that each individual new memory (including modified memories generated for an overwrite/merge) would go through an additional temporal post-processing step to make memories with specific dates easier for the model to understand. Our tiny, quantized Qwen model often has issues reasoning over dates, such that even if we tell Qwen when a memory was extracted and the current date, it may respond to the user as if the event hasn't happened yet even though the event has since passed. For example, it might say \emph{`Remember, you have a test on Wednesday'} even though Wednesday has passed and it is now Thursday. To minimize the amount of reasoning our model needs to do over dates, the post-processing step takes a memory such as \emph{`The child has a test this Wednesday.'} and given today's date and a calendar of 90 days centered around the current date, would convert it to a code-interpretable date format, for example: \emph{`The child has/had a test on \{10-08-2025\}'}. Then during retrieval if we notice braces (\{\}) in a retrieved memory, via a set of code rules we can substitute in an appropriate more interpretable version of the date. For example, a conversation on the 10th would be \emph{`The child has/had a test on this last Wednesday, 10/08/2025.'}. While the resulting sentence might not be grammatically correct, Qwen is a lot better at understanding that than reasoning over numerical dates.

Where the tiny qwen model ultimately struggled was at the step of making the reasoning from a statement like \emph{`this Wednesday'} to the exact date based on a small calendar, with the current date specified. Instead, it would make the same types of mistakes we were hoping to fix, where it would use the date of a different Wednesday instead. 

\subsubsection{Why would we want to post-process}
To give a more comprehensive example of why we would ideally like to post-process our memories, lets imagine that today's date is Sunday, October 5th 2025. If the user says something like \emph{`I have a test on Wednesday'}, it is understood to us that Wednesday in this context refers to Wednesday October 8th 2025. Now the memory extraction model might extract the memory \emph{`The user has a test on Wednesday'}, and our system will know that this memory was extracted on October 5th 2025. Now lets imagine the user has another conversation on Thursday November 10th 2025. Since the test date has passed, the system should ideally say something like \emph{`How was your test yesterday?'}. However, we found that our tiny quantized model of Qwen cannot reason that the date has passed, even when told that \emph{`The Date today is Thursday 10/9/2025.'} with the memory inserted as \emph{`The child has a test this Wednesday. {-}- Memory Extracted on 10/5/2025'}. Here's an example output from this test: if the user enters: \emph{`hi'}, the assistant responded with: \emph{`Hi there! How are you feeling today? Remember, you have a test on Wednesday. You've been studying hard, so I'm sure you'll do great!'.} 
\subsubsection{Implementation Code Rules Details}
Given an example post-processed memory: \emph{`The child has/had a test on {10-05-2025}'}. A conversation on the 6th, would pull that memory as \emph{`The child has/had a test on this upcoming Wednesday, 10/8/2025.'}, a conversation on the 9th would be \emph{`The child has/had a test on yesterday Wednesday, 10/8/2025.'} and a conversation on the 10th would be \emph{`The child has/had a test on this last Wednesday, 10/8/2025.'}. For weeks or months, the appropriate text would be something like `[upcoming/past] Wednesday, 10/8/2025, a [few weeks/month/few months] [from now/ ago]' depending on whether its in the past or future. 

\section{Our Hyperparameter Settings}
We have not performed rigorous hyperparameter tuning nor comparison, and consequently do not claim that these are necessarily the optimal values for our use case. Naturally, given a different use case, different hyperparameter values might be ideal.
\begin{itemize}[label={}]
\item top-$k_\text{short} = 3$
\item top-$k_\text{long} = 5$
\item $S_\text{min} = .35$ 
\item $W_\text{slide} = 15$
\item $k_\text{surrounding}= 1$
\item $T_\text{inactivity} = 10$ minutes
\item $c_\text{chunk} = 7000$ tokens
\item $k_\text{overlap} = 5$
\item $R_\text{min} = 0.01$
\end{itemize}

\section{Code}
Due to commercial constraints, we do not plan to release the full codebase associated with this project, especially code related to our specific system. However, we do plan to release code associated with our benchmark and evaluation process at a later date.

\end{document}